\documentclass[10pt,twocolumn,letterpaper]{article}

\usepackage{cvpr}              
\usepackage{subcaption}
\usepackage{multicol}
\usepackage{array}
\usepackage{multirow}
\usepackage{algorithm}
\usepackage{algpseudocode}

\algnewcommand\Algphase[1]{\vspace{0.5em}\State \textbf{#1}}

\definecolor{cvprblue}{rgb}{0.21,0.49,0.74}
\usepackage[pagebackref,breaklinks,colorlinks,allcolors=cvprblue]{hyperref}

\def\paperID{*****} 
\def\confName{CVPR}
\def\confYear{2026}

\title{Anatomica: Localized Control over Geometric and Topological Properties for Anatomical Diffusion Models}

\author{
    Karim Kadry$^{1}$\footnotemark[1] \quad
    Abdalla Abdelwahed$^{2}$\footnotemark[1] \quad
    Shoaib Goraya$^{3}$ \quad
    Ajay Manicka$^{1}$ \\
    Naravich Chutisilp$^{4}$ \quad
    Farhad R.~Nezami$^{3}$ \quad
    Elazer R.~Edelman$^{1}$ \\[0.35em]
    {\small $^{1}$MIT, Cambridge, MA, USA \quad
            $^{2}$American University in Cairo, New Cairo, Egypt} \\
    {\small $^{3}$Brigham and Women’s Hospital, Boston, MA, USA \quad
            $^{4}$EPFL, Lausanne, Switzerland}
}

\begin{document}
\twocolumn[{%
\renewcommand\twocolumn[1][]{#1}%
\maketitle
\centering
\includegraphics[width=0.9\textwidth]{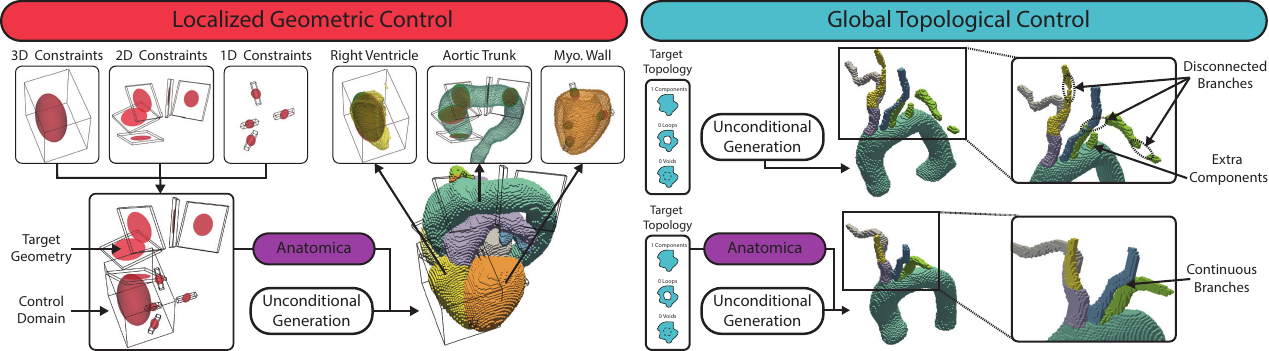}
\captionof{figure}{\textbf{Anatomica is a compositional diffusion-guidance framework for generating segmentations based on anatomical features that are localized within cuboidal control domains.} \textbf{Left}: We generate voxel maps according to localized target geometry (size, shape, and position) visualized as red ellipsoids. \textbf{Right}: We generate voxel maps according to target topology (components, loops, and voids).
}

\label{fig:teaser}
\vspace{1em}
}]

\begingroup
\renewcommand\thefootnote{\fnsymbol{footnote}}%
\footnotetext[1]{Equal contribution.}%
\endgroup

\begin{abstract}
We present Anatomica: an inference-time framework for generating multi-class anatomical voxel maps with localized geo-topological control. During generation, we use cuboidal control domains of varying dimensionality, location, and shape, to slice out relevant substructures. These local substructures are used to compute differentiable penalty functions that steer the sample towards target constraints. We control geometric features such as size, shape, and position through voxel-wise moments, while topological features such as connected components, loops, and voids are enforced through persistent homology. Lastly, we implement Anatomica for latent diffusion models, where neural field decoders partially extract substructures, enabling the efficient control of anatomical properties. Anatomica applies flexibly across diverse anatomical systems, composing constraints to control complex structures over arbitrary dimensions and coordinate systems, thereby enabling the rational design of synthetic datasets for virtual trials or machine learning workflows.
\end{abstract}    
\section{Introduction}
\label{sec:intro}
Anatomical form plays a vital role in dictating the function and dysfunction of physiological systems. By virtually modelling patient-specific organ systems as 3D voxelized segmentations, we can leverage numerical simulators to reveal structure-function relationships that inform clinical research and medical device design. Such use cases include the simulation of clinical trials to evaluate medical devices \cite{sarrami2021silico,viceconti2021possibleinsilico,abadi2020virtual}, or simulating image-formation to create robust datasets for machine learning workflows \cite{gopalakrishnan2024intraoperative,gopalakrishnan2022fast,dey2024learning,billot2023synthseg,fernandez2024generating}.

Due to the sparsity and imbalances inherent to real-world datasets, there has been growing interest in augmenting anatomical datasets with synthetic data. A key advantage of using generative models over patient datasets lies in their controllability. Conditional generation of medical images based on anatomical or demographic information has been shown to improve the performance of machine learning classifiers and segmentation networks \cite{ktena2024generative,moroianu2025improving,fernandez2024generating}. However, conditional generation of 3D multi-class segmentations based on anatomical features remains difficult. These features encompass both \textit{geometry} (shape and size) and \textit{topology} (connected components, loops, or voids). Moreover, such features are defined compositionally over multiple substructures within the segmentation, with varying dimensionality (e.g., 3D vs 2D), and across varying coordinate systems (e.g., Cartesian vs curvilinear). The ideal generative model must not only control such features in a \textit{precise} and \textit{compositional} manner, but also offer control mechanisms that are \textit{intuitive} to use.

We introduce Anatomica: an inference-time framework for controlling anatomical latent diffusion models based on arbitrarily localized properties related to geometry and topology. We formulate guidance through two key stages for each sampling step. First, we differentiably parse voxel-space segmentations to extract anatomical substructures with varying dimensionality over arbitrary coordinate systems. Second, we measure geometric and topological properties in a differentiable manner and apply potential functions to guide the reverse sampling process. Lastly, we adapt this guidance framework for latent diffusion models through neural field decoders which map arbitrary query points in latent space to voxel space, enabling the efficient measurement of anatomical properties from latent space. We advance the state-of-the-art in the following ways:
\begin{itemize}
    \item \textbf{Differentiable and Localized Substructure Extraction}: We introduce a modular method to differentiably parse \textit{localized} and \textit{anatomically relevant} substructures from voxel-space segmentations (V-parsing). We base our method on cuboidal control domains with varying scales, positions, and orientations. By arranging multiple control domains of varying dimensionality across relevant coordinate systems, we enable the characterization of a wide array of anatomical systems and structures.
    \item \textbf{Unified Geo-Topological Measurement and Guidance}: We demonstrate that applying differentiable measurement and potential functions over anatomical substructures allows us to constrain localized properties through diffusion guidance. This includes geometric properties such as size, shape, position, and orientation, as well as topological properties such as the number of components, loops, or voids. We show that, by combining different control domains and potential, we unlock a rich design space for \textit{compositional} anatomical control, within which a wide variety of structures can be controllably generated.
    \item \textbf{Latent Diffusion Guidance with Neural fields}: We show that neural field decoders enable the efficient measurement of voxel-space properties within control domains \textit{directly from latent space} during sampling (L-parsing). By exploiting the ability of neural fields to decode arbitrarily discretized point grids, we avoid the computational overhead of full-volume decoding. We introduce two partial decoding strategies: \textit{coarse L-parsing} decodes globally at reduced spatial resolution, while \textit{localized L-parsing} decodes local regions at high resolution.
\end{itemize}

\section{Related Work}
\textbf{Geometric Control for Generative Models of Anatomy}
Geometric features such as size and shape play a crucial role in biophysical dynamics \cite{fabris2022thinsmall,kadry2021platform}. Modelling anatomy with simple shapes such as cylinders \cite{arostica2025software,madani2019bridgingfemdeeplearningstress} provides control over form but not realism. Statistical shape models \cite{dou2022generativechimeras,qiao2025personalized,williams2022aortic} represent realistic variation via global shape vectors \cite{hermida2024onsetdeviccistatisticalshape,williams2022aortic} but are not as interpretable or editable. To bridge this gap, recent studies conditionally train generative models based on size-based measures \cite{de2025steerable,kadry2024morphology}. Recently, \citet{kadry2025cardiocomposer} proposed inference-time geometric guidance via differentiable geometry, expanding control to size, position, and shape, in a compositional manner over multi-class anatomy. However, this method was limited to globally defined geometric properties in 3D. In this work, we extend geometric guidance to arbitrarily localized attributes based on cuboidal control domains of varying scale, position, orientation, and dimensionality. By arranging control domains over non-Cartesian coordinate systems, we enable significantly more complex compositional control of physiologically relevant geometric features.
\begin{figure*}[t]
    \centering
    \includegraphics[width=\textwidth]{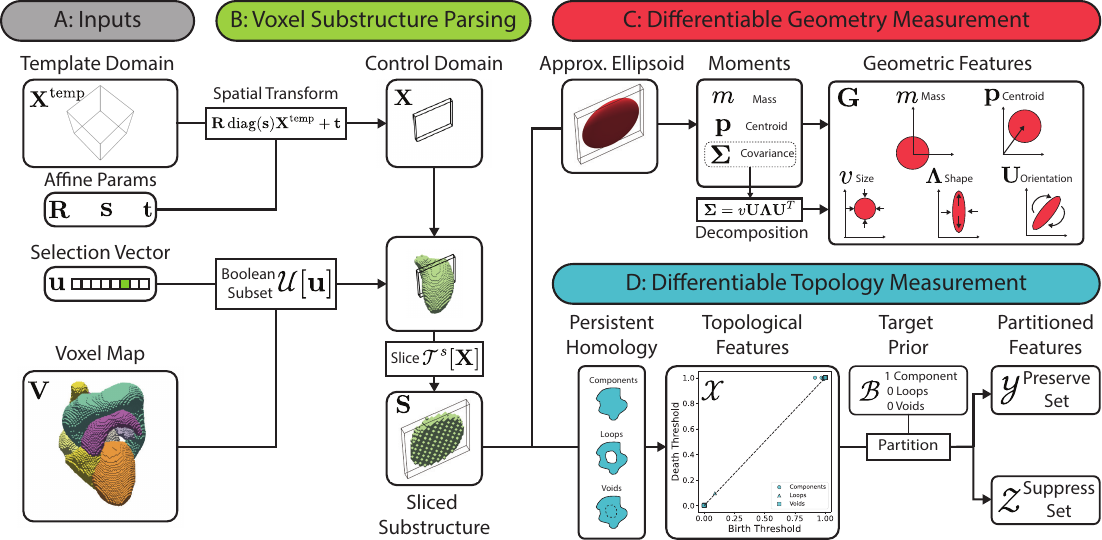}
    \caption{
        \textbf{Differentiable measurement of anatomical properties from multi-class voxel maps.} \textbf{A}: We differentiably parse relevant substructures from anatomical voxel maps for localized measurement. \textbf{B}: We spatially transform cuboidal primitives (template domains) into control domains that slice into anatomical structures (V-parsing). \textbf{C}: The substructure is then differentiably measured in terms of geometric properties; as well as \textbf{D}: persistent homology-based topological properties.
    }
    \label{fig:substructure_parsing}
\end{figure*}

\noindent\textbf{Topological Deep Learning} 
Topological properties such as the number of components, loops, or voids also play a crucial role in modulating biophysical dynamics \cite{kong2024sdf4chd}. To regularize machine learning workflows in a differentiable manner, persistent homology (PH) can be used \cite{bruel2019topology} to measure the continuous-valued persistence of topological features. PH-based topological losses have been used for the training \cite{clough2020topologicalloss} and test-time adaptation \cite{byrne2022persistenthomologymulticlass} of segmentation networks. Similarly, PH has been used to conditionally train diffusion models of 2D binary label maps \cite{gupta2025topodiffusionnet} and 3D surfaces \cite{hu2024topology}. In contrast to using topological losses to update network weights, we use PH for inference-time control generative models that sample multi-class 3D anatomical segmentations without conditional training. This enables us to flexibly constrain topological features in a plug-and-play manner without retraining.

\noindent\textbf{Spatial Conditioning for Generative Models}
Spatial control of generative models relies on two main strategies. The first conditions models on mid-level representations (e.g., bounding boxes, ellipsoid parameters) \cite{nie2024compositionalblobgen,feng2025blobgen,hertz2022spaghetti,koo2023salad,anonymous2024protcomposer}. The second involves guidance methods, such as self-guidance \cite{epstein2023diffusionselfguidance}, which employs attention-based losses for basic geometric control (size, position) in text-to-image models, but it is not suited for multi-label segmentations, nor is it adapted for complex constraints needed to describe anatomical shape. In our work, we extend energy-based guidance to localized control over geometry and topology by introducing differentiable potentials for 3D multi-component anatomical voxel maps based on substructure-specific properties. We show that this enables a rich design space for anatomical control, within which a wide variety of organs can be controllably generated.
\section{Methodology}

\subsection{Anatomical Latent Diffusion Models}
\label{sec:methods_diffusion}
\textbf{Autoencoder with Neural Field Decoder} We develop our variational autoencoder based on hybrid implicit-explicit representations \cite{peng2020convolutional}. Our dataset consists of 3D segmentation volumes $\mathbf{V} \in \mathbb{R}^{C \times H \times W \times D}$ with $C$ tissue channels and $(H, W, D)$ spatial dimensions. During training, a convolutional encoder $\mathcal{E}$ first encodes the voxelized segmentation map $\mathbf{V}$ into a voxelized latent grid representation $\mathbf{z} = \mathcal{E}(\mathbf{V})$, where $\mathbf{z} \in \mathbb{R}^{c \times h \times w \times d}$ comprises $c$ channels and spatial dimensions $(h,w,d)=(H/f,\,W/f,\,D/f)$ for an integer downsampling factor $f$. To decode back into voxel space, a 3D query point grid $\mathbf{X}^q \in \mathbb{R}^{H \times W \times D \times 3}$ is used to compute a latent point grid through the latent slice operator $\mathcal{T}^l[\mathbf{X}^q]: \mathbb{R}^3 \to \mathbb{R}^c$ which applies trilinear interpolation for each query point as in \citet{jaderberg2015spatial}. The resulting latent grid is then pointwise decoded with a neural field decoder $\mathcal{F}: \mathbb{R}^c \to \mathbb{R}^C$ into the predicted voxel map $\bar{\mathbf{V}} \in \mathbb{R}^{C \times H \times W \times D}$:

\begin{figure*}[h]
    \centering
    \includegraphics[width=\textwidth]{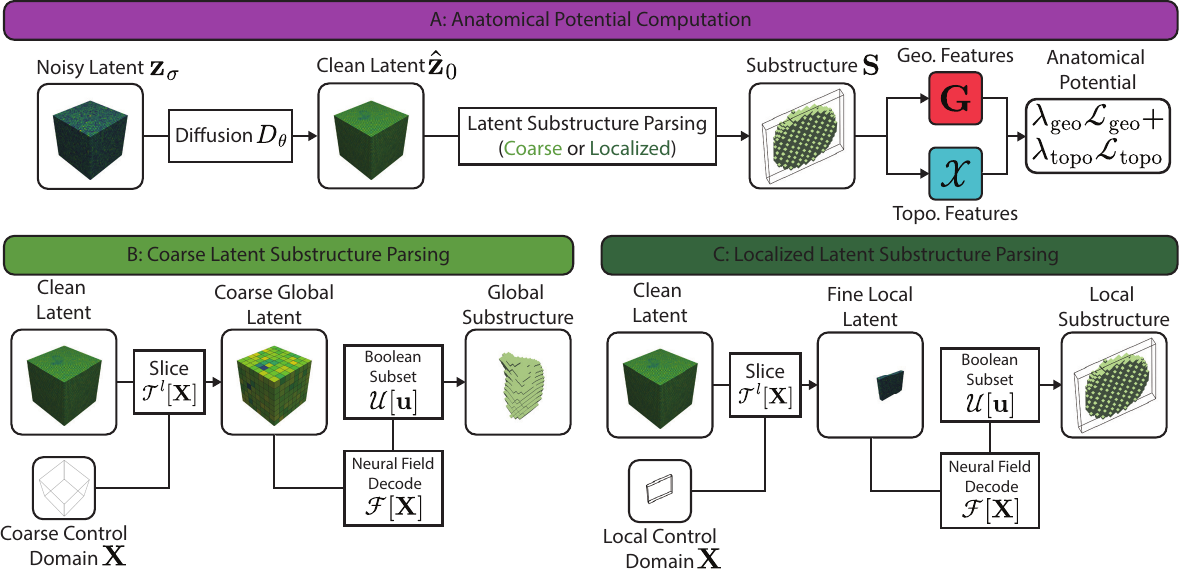}
    \caption{
        \textbf{Efficient parsing of anatomical substructures during diffusion guidance.}
        \textbf{A}: During guidance, we parse relevant substructures directly from the clean latent prediction with a neural field decoder (L-parsing). \textbf{B}: In coarse L-parsing, we use a coarse grid to decode globally defined substructures at low spatial resolution. \textbf{C}: In localized L-parsing, we use a similar grid size but spatially transform the template point grid to decode localized substructures at high spatial resolution. 
    }
    \label{fig:partial_decoding_guidance}
\end{figure*}
\begin{equation}
    \bar{\mathbf{V}} = \underbrace{\mathcal{F}[\mathbf{X}^q]}_{\text{Decode Latent}} \circ \underbrace{\mathcal{T}^l[\mathbf{X}^q](\mathbf{z})}_{\text{Slice Latent}}.
\end{equation}
Here, $\mathcal{F}$ is parametrized as a multi-layer perceptron that takes in the interpolated latent point grid and positionally encoded query points, and $\circ$ denotes function composition.

\noindent\textbf{Unconditional Diffusion Model} We use an unconditional latent diffusion model (LDM) as a prior over 3D anatomical segmentations. In the forward process, data samples are progressively corrupted by adding Gaussian noise $\mathbf{n}$ through the relation $\mathbf{z}_\sigma = \mathbf{z} + \mathbf{n}$ where $\mathbf{n} \sim \mathcal{N}(\mathbf{0}, \sigma^2 \mathbf{I})$. Similar to \citet{karras2022elucidating}, we aim to learn the score function $\nabla_{\mathbf{z}_\sigma} \log p(\mathbf{z}_\sigma; \sigma)$ that defines the reverse diffusion process:
\begin{equation}
d\mathbf{z}_\sigma = -2\sigma \nabla_{\mathbf{z}_\sigma} \log p(\mathbf{z}_\sigma; \sigma) \, dt + \sqrt{2\sigma}\, d\mathbf{w} \,
\end{equation}
where $d\mathbf{w}$ is the Wiener process. This score function $\nabla_{\mathbf{z}_\sigma} \log p(\mathbf{z}_\sigma; \sigma) = (D_\theta(\mathbf{z}_\sigma; \sigma) - \mathbf{z}_\sigma)/\sigma^2$ can be expressed via a denoising function $D_\theta$ parametrized by a 3D U-Net. The neural network is trained by minimizing the clean data prediction objective $L = \mathbb{E}_{\sigma, \mathbf{z}, \mathbf{n}} \left[ \omega(\sigma) \| D_\theta(\mathbf{z}_\sigma; \sigma) - \mathbf{z} \|_2^2 \right]$, with $\omega(\sigma)$ balancing loss contributions across noise levels. 

\subsection{Guidance via Anatomical Potential Functions}
\label{sec:methods_guidance}
We assume the voxel map $\mathbf{V}$ contains $K$ substructures $\mathbf{S}_k \in [0,1]^{\alpha \times \beta \times \gamma}$ of interest, where $\alpha, \beta, \gamma$ are grid size dimensions. Substructures represent anatomical regions of interest that may be comprised of single tissues (e.g., cross-sectional slices of an aorta) or combinations thereof (e.g., left and right atria). As shown in \cref{fig:substructure_parsing}, each substructure can be measured in terms of geometric properties (mass, position, size, shape, orientation) and topological properties (connectivity, presence of loops or voids).

\noindent\textbf{Differentiable Geometric Measurement} Given a single parsed substructure $\mathbf{S}_k$, we aim to differentiably extract the geometric moments by numerically integrating the zeroth, first, and second-order moments of the substructure. Here, the zeroth moment represents the mass $m_k \in \mathbb{R}$, the first moment represents the centroid $\mathbf{p}_k \in \mathbb{R}^3$, and the second moment represents the covariance matrix $\boldsymbol{\Sigma}_k \in \mathbb{R}^{3\times 3}$. To do this, we follow \citet{kadry2025cardiocomposer}, but instead compute the moments for substructures inside arbitrarily sized cuboidal control domains, rather than globally. We first define $\mathbf{\Omega}_k \in \mathbb{R}^{(\alpha\beta\gamma)\times 1}$ as the flattened substructure voxel grid $\mathbf{S}_k$ and $\mathbf{r}_k \in \mathbb{R}^{(\alpha\beta\gamma)\times 3}$ as the normalized voxel coordinates between 0 and 1. The moments are then computed as:
\begin{align}
    \label{eq:geometric_moments}
m_k  &= \mathbf{1}^T \cdot \mathbf{\Omega}_k 
\quad\mathrm{and}\quad 
\mathbf{p}_k  = \frac{\mathbf{\Omega}_k^T \mathbf{r}_k}{m_k} \nonumber,\\
\boldsymbol{\Sigma}_k &= \frac{1}{m_k} \mathbf{r}_k^T \operatorname{diag}(\mathbf{\Omega}_k)\,\mathbf{r}_k - \mathbf{p}_k \mathbf{p}_k^T.
\end{align}
Here, $\mathbf{1}^T$ is the all-ones vector, and $\operatorname{diag}(\cdot)$ refers to diagonal matrix embedding.

\noindent\textbf{Geometric Decomposition and Size Normalization} As the covariance matrix implicitly contains information on mass, we aim to obtain a scale-normalized covariance matrix that relates only to orientation and relative aspect ratios. We decompose the covariance as $\boldsymbol{\Sigma}_k = \mathbf{U}_k\tilde{\boldsymbol{\Lambda}}_k \mathbf{U}_k^T = v_k\mathbf{U}_k \boldsymbol{\Lambda}_k \mathbf{U}_k^T$, where size $v_k \in \mathbb{R}$ = $\operatorname{tr}(\boldsymbol{\Sigma}_k)$ is the trace of the covariance matrix, shape $\boldsymbol{\Lambda}_k \in \mathbb{R}^{3 \times 3}$ = $\tilde{\boldsymbol{\Lambda}}_k / v_k$ is the eigenvalue matrix $\boldsymbol{\Lambda}_k$ normalized by the trace, and orientation $\mathbf{U}_k \in \mathbb{R}^{3 \times 3}$ is an orthonormal matrix. We thus define the scale-normalized covariance matrix as $\boldsymbol{\Sigma}^n_k = \mathbf{U}_k\boldsymbol{\Lambda}_k \mathbf{U}_k^T$.

\noindent\textbf{Geometric Potential Functions} After computing the geometric features, we aim to penalize the deviations from the target geometric features $\bar{\mathbf{G}}_k$ through a geometric potential function $\mathcal{L}^{\text{geo}}_k$. We formulate this potential function as a weighted combination of mean squared error losses $\mathcal{L}_{\text{MSE}}$:

\begin{align}
    \mathcal{L}^{\text{geo}}_k = \lambda_0\mathcal{L}_{\text{MSE}}(m_k, \bar{m}_k) + \lambda_1\mathcal{L}_{\text{MSE}}(\mathbf{p}_k, \bar{\mathbf{p}}_k) \nonumber\\ + \lambda_2 \mathcal{L}_{\text{MSE}}(\boldsymbol{\Sigma}^n_k, \bar{\boldsymbol{\Sigma}}^n_k).
\end{align}
Here, $[\lambda_0, \lambda_1, \lambda_2]$ are the weighting factors for the mass, position, and normalized covariance losses, respectively.

\noindent\textbf{Adaptive Mass Weighting} Given that geometric features are now defined locally, rather than globally, our potential functions must account for numerical instability in the centroid and covariance-based losses resulting from empty voxels. We address this by adaptively setting $\lambda_1 = \lambda_2 = 0$ when the mass is below a specified threshold.

\noindent\textbf{Differentiable Topological Measurement}
We use persistent homology (PH) to measure the presence of topological features within the substructure $\mathbf{S}_k$ considered as a cubical complex, similar to \citet{gupta2025topodiffusionnet}. We consider super-level sets of $\mathbf{S}_k$, which are the set of voxels above a threshold value $\tau$. By decreasing $\tau$, we obtain a filtration of nested sets, all of which are used to extract topological features such as connected components (0D features), loops (1D features), and voids (2D features). The output of this process is a set of persistence points $p \in \mathcal{X}_k$ which include the birth $b$ and death $d$ thresholds for all topological features. Intuitively, persistent features have a large interval between their birth and death thresholds. We use the Cubical Ripser library \cite{kaji2020cubical} to compute the PH of $\mathbf{S}_k$.

\noindent\textbf{Softmax Temperature Tuning} As our substructure is derived from softmaxed multi-class probability maps, we found that the gradient of the topological potential was too low in regions where the probability was close to 0 or 1. To address this, when decoding the substructure from the latent space, we apply softmax with an increased temperature for topological guidance, enabling the gradient to pass through during backpropagation. 

\noindent\textbf{Topological Potential Functions}
To enforce topological structures within $\mathbf{S}_k$  during the reverse diffusion process, we partition the persistence set into disjoint sets $\mathcal{X}_k = \mathcal{Y}_k \cup \mathcal{Z}_k$ consisting of points that should be preserved $\mathcal{Y}_k$ or suppressed $\mathcal{Z}_k$ based on a topological prior $\mathcal{B}_k \in \mathbb{R}^{3}$, which specifies the desired features for the components, loops, and voids, respectively. To differentiably compute the topological potential, we sample the voxel intensities from $\mathbf{S}_k$ at the birth $r^p_b$ and death $r^p_d$ coordinates for each persistence point $p$. We then maximize or minimize the persistence of each feature through the following potential functions:
\begin{figure}[h]
    \centering
    \includegraphics[width=\columnwidth]{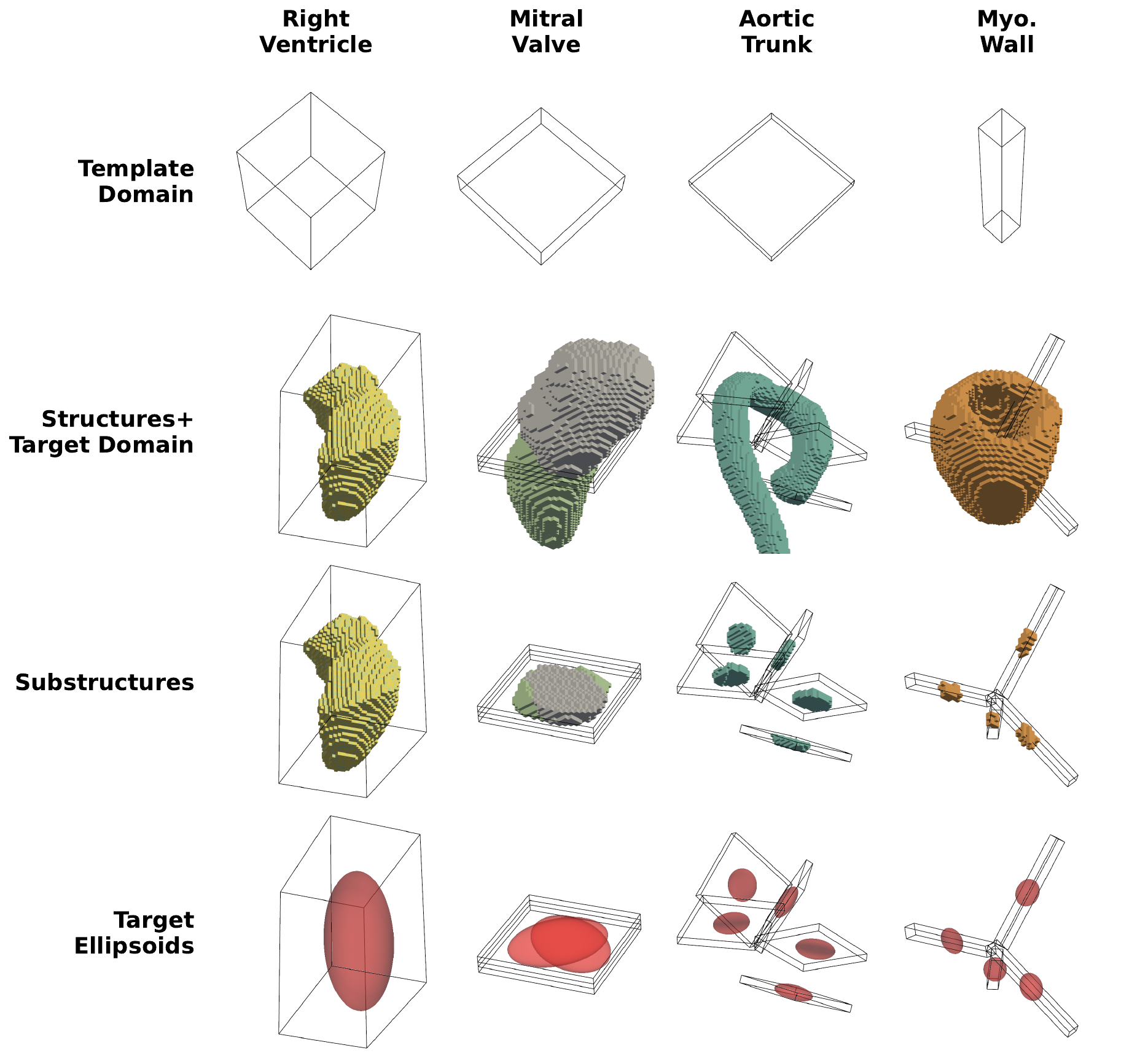}
    \caption{
        \textbf{Geometric Control Tasks}. We define a variety of relevant tasks by varying the selected tissues, template domain grid size, and control domain-specific spatial transforms.
    }
    \label{fig:design_space}
\end{figure}
\begin{equation}
    \scalebox{0.75}{ 
    $\mathcal{L}^{\text{topo}}_k = -\sum_{p \in \mathcal{Y}_k} \underbrace{|\mathbf{S}_k(r^p_b) - \mathbf{S}_k(r^p_d)|^2}_{\text{Preserve Loss}} + \sum_{p \in \mathcal{Z}_k} \underbrace{|\mathbf{S}_k(r^p_b) - \mathbf{S}_k(r^p_d)|^2}_{\text{Suppress Loss}}$
    }.
\end{equation}

\noindent\textbf{Gradient-Based Guidance} Following \citet{kadry2025cardiocomposer}, we guide the diffusion process using the gradient derived from anatomical potential functions. At each sampling step, we first denoise the intermediately noised latent $\mathbf{z}_\sigma$ to obtain a clean latent prediction $\hat{\mathbf{z}}_0=D_{\theta}(\mathbf{z}_\sigma;\sigma)$, which is then \textit{parsed} into $K$ substructures $\mathbf{S}_k$ as described in \cref{sec:parsing}. We compute a composite anatomical potential $\mathcal{L}=\frac{1}{K}\sum_{k} (\lambda_{\text{geo}}\mathcal{L}^{\text{geo}}_k + \lambda_{\text{topo}}\mathcal{L}^{\text{topo}}_k)$ where $\lambda_{\text{geo}}$ and $\lambda_{\text{topo}}$ weight the geometric and topological potentials (\cref{fig:partial_decoding_guidance}, top row). The guided denoising step is then applied as follows:
\begin{equation}
    \scalebox{0.9}{ 
    $\underbrace{D_{\theta}^w (\mathbf{z}_\sigma;\sigma)}_{\text{Guided Denoising}} = \underbrace{D_{\theta}(\mathbf{z}_\sigma;\sigma)}_{\text{Uncond.\  Denoising}}-\underbrace{\sigma^2 \cdot \nabla_{\mathbf{z}_\sigma}\mathcal{L}}_{\text{Anatomical Guidance}}$.
    }
\end{equation}

\subsection{Substructure Parsing for Diffusion Guidance}
\label{sec:parsing}
\noindent\textbf{Voxel-Space Parsing} Given a predicted voxel-space segmentation map $\hat{\mathbf{V}}$, we can parse $K$ substructures $\mathbf{S}_k$ with voxel-space substructure parsing (\cref{fig:substructure_parsing}), which we will refer to as \textit{V-parsing}. V-parsing extracts a substructure by first using the boolean subset operator $\mathcal{U}[\mathbf{u}] : \mathbb{R}^C \to \mathbb{R}$; parametrized by a selection vector $\mathbf{u} \in \{0,1\}^C$ for elementwise extraction and recombination of tissues from $\hat{\mathbf{V}}$ into a structure voxel map $\hat{\mathbf{S}}_k$. By taking different tissue combinations, we enable the control of various structures within the segmentation. The structure slicing operator $\mathcal{T}^s[\mathbf{X}_k] : \mathbb{R} \to \mathbb{R}$ then samples voxelwise intensity values from $\hat{\mathbf{S}}_k$ at locations specified by a cuboidal control domain $\mathbf{X}_k \in \mathbb{R}^{\alpha \times \beta \times \gamma \times 3}$ discretized into a lattice-like point grid. The final relation is given by:
\begin{equation}
    \mathbf{S}_k = \underbrace{\mathcal{T}^s[\mathbf{X}_k]}_{\text{Slice Structure}} \circ \underbrace{\mathcal{U}[\mathbf{u}](\hat{\mathbf{V}})}_{\text{Subset Voxel}}.
\end{equation}

\noindent\textbf{Spatial Transformation of Template Domains} Control domains $\mathbf{X}_k$ are obtained by spatially transforming a template point grid $\mathbf{X}^{\text{temp}}_k \in \mathbb{R}^{\alpha \times \beta \times \gamma \times 3}$ using affine transformation parameters $\mathbf{A}_k=[\mathbf{R}_k, \mathbf{s}_k, \mathbf{t}_k]$, where $\mathbf{R}_k \in \mathbb{R}^{3 \times 3}$ is a rotation matrix, $\mathbf{s}_k \in \mathbb{R}^{3}$ is a scaling vector, and $\mathbf{t}_k \in \mathbb{R}^3$ is a translation vector. The spatial transformation is defined as:
\begin{equation}
    \mathbf{X}_k= \mathbf{R}_k\operatorname{diag}(\mathbf{s}_k)\mathbf{X}^{\text{temp}}_k + \mathbf{t}_k.
\end{equation}
This formulation enables flexible substructure parsing across multiple design axes (\cref{fig:design_space}). For example, the template grid size controls the discretization (coarse to fine) and dimensionality (3D to 1D) of the parsed substructure. The affine parameters $\mathbf{A}_k$ can be defined based on custom coordinate systems and enable the localization of substructures under varying extents, orientations, and positions in 3D space.

\noindent\textbf{Latent-Space Parsing} Rather than decoding the entire voxel grid and subsequently slicing out substructures, we propose to use neural field decoders to directly parse substructures from latent space (\textit{L-parsing}). This is accomplished by applying the latent slicing operator $\mathcal{T}^l[\mathbf{X}_k]$ on the clean predicted latents $\hat{\mathbf{z}}_0$ followed by the neural field decoder $\mathcal{F}[\mathbf{X}_k]$ and the boolean subset operator $\mathcal{U}[\mathbf{u}]$:
\begin{equation}
    \mathbf{S}_k = \underbrace{\mathcal{U}[\mathbf{u}]}_{\text{Subset Voxel}} \circ \underbrace{\mathcal{F}[\mathbf{X}_k]}_{\text{Decode Latent}} \circ \underbrace{\mathcal{T}^l[\mathbf{X}_k]( \hat{\mathbf{z}}_0)}_{\text{Slice Latent}}.
\end{equation}
\noindent\textbf{Partial Decoding Strategies} Neural field representations offer a key advantage in that they enable decoding from arbitrary point sets. We introduce two L-parsing strategies that apply \textit{partial decoding} by using point grids with small grid sizes, enabling fast guidance at inference time. To efficiently measure global-level anatomical properties that are invariant to spatial resolution, we apply \textbf{coarse L-parsing} by using a template grid $\mathbf{X}^\text{temp}_k$ with a low grid discretization $[\alpha<H, \beta<W, \gamma<D]$ and setting the identity-like affine transformation parameters $\mathbf{A}^{\text{coarse}}=[\mathbf{I}, \mathbf{1}, \mathbf{0}]$ (\cref{fig:partial_decoding_guidance} bottom left). To efficiently measure anatomical properties that are defined locally, we apply \textbf{localized L-parsing} by using a similarly low template grid discretization, but spatially transform the template point grid $\mathbf{X}^\text{temp}_k$ into a localized region, effectively achieving a high spatial resolution (\cref{fig:partial_decoding_guidance} bottom right).

\section{Experiments}
\subsection{Anatomical Datasets}
For the \textbf{cardiac dataset}, we extract heart-related labels from the TotalSegmentator dataset \cite{wasserthal2022totalsegmentator}, resulting in 596 11-channel segmentations. For the \textbf{aortic dataset}, we extract aorta-related labels from the TotalSegmentator dataset, resulting in 450 7-channel segmentations. For the \textbf{spinal dataset}, we extract spinal vertebrae-related labels from the CTSpine1k dataset \cite{deng2021ctspine1k}, resulting in 784 25-channel segmentations. For the \textbf{coronary dataset}, we extract coronary artery-related labels from the DISRUPT-CAD dataset \cite{visinoni2024coronary}, resulting in approximately 360 unique 4-channel segmentations. All segmentations were resampled to a uniform resolution of $128^3$, and were partitioned into training and validation sets with an 80/20 split.

\subsection{Control Tasks}
\noindent\textbf{Geometric Control} To evaluate geometric control, we use the cardiac validation set to determine appropriate target control domains and target geometric features (see \cref{fig:design_space}). The first task is \textbf{Right Ventricle}, which constrains the right ventricle within a volumetric domain. The second task is \textbf{Mitral Valve}, which constrains two volumetric domains at the intersection of the left ventricle and atrium. The third task is \textbf{Aortic Trunk}, which constrains 5 planar domains along the aortic trunk centerline. The fourth task is \textbf{Myocardium Wall}, which constrains 4 linear domains emanating from the centroid of the myocardium. All quantitative experiments were conducted by sampling 128 synthetic samples.

\noindent\textbf{Topological Control} To evaluate topological control, we define a task for each anatomical dataset. We set a global target control domain as well as appropriate topological priors for each task: \textbf{Atria Separation} enforces 2 connected components for the left and right atria; \textbf{Branch Connectivity} enforces 1 connected component for the ascending aorta and all branches; \textbf{Vertebrae Connectivity} enforces 1 connected component and 9 loops for five thoracic spinal vertebrae (T6-T10); and \textbf{Calcium Count} enforces 2 calcium components in the coronary artery wall.
\renewcommand{\arraystretch}{1.25}
\begin{table}[h]
\centering
\caption{\textbf{Comparison of geometric control task approaches}.}
\label{tab:B0}
\resizebox{\columnwidth}{!}{\begin{tabular}{cccc}
\toprule
\textbf{Approach} & \textbf{Decoder} & \textbf{Parsing} & \textbf{Control Method} \\
\midrule
Explicit & Neural Field & N/A & Conditioning \\
Implicit & Neural Field & N/A & Conditioning \\
Anatomica-V & Convolutional & V-parsing & Guidance \\
Anatomica-L & Neural Field & L-parsing & Guidance \\
\bottomrule
\end{tabular}
}
\end{table}
 
\begin{figure}[t]
    \centering
    \includegraphics[width=\columnwidth]{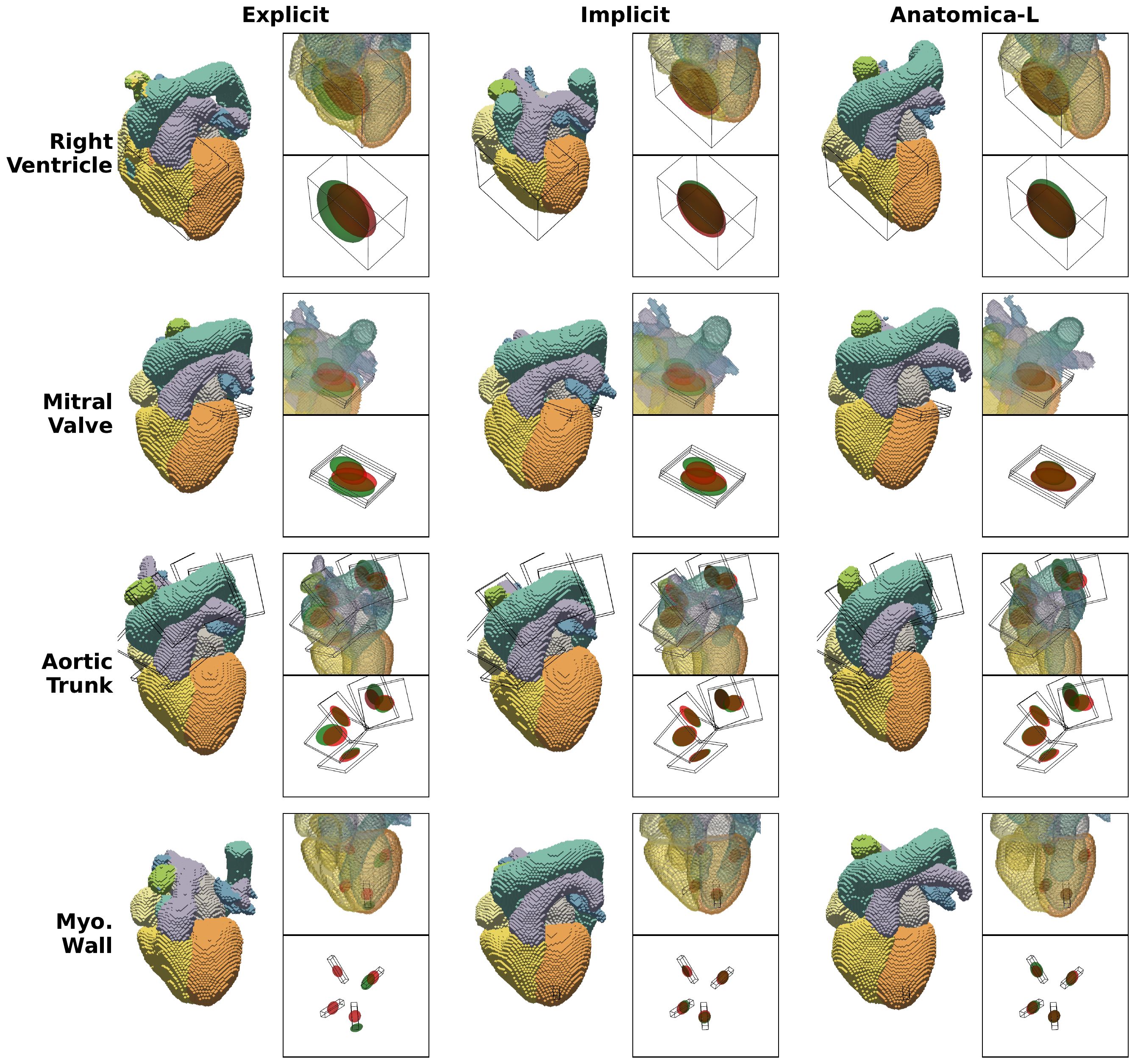}
    \caption{
        \textbf{Qualitative evaluation of geometric control experiments.} We generate anatomical segmentations based on target domains (black frames) and geometric features (red ellipsoids) for four cardiac tasks. Sample geometry shown as green ellipsoids.
    }
    \label{fig:geo_control}
\end{figure}
\renewcommand{\arraystretch}{1.25}
\begin{table}[h]
\centering
\caption{\textbf{Quantitative results for geometric control tasks}. We report geometric fidelity and generation quality for each task-approach combination. Fidelity values for mass, centroid, and covariance are multiplied by 1e5, 1e4, 1e5 respectively.}
\label{tab:B1}
\resizebox{0.9\columnwidth}{!}{\begin{tabular}{ccccccc}
\toprule
 &  & \multicolumn{3}{c}{\textbf{Geometric Fidelity (↓)}} & \multicolumn{2}{c}{\textbf{Generation Quality (↓)}} \\
 \cmidrule(r){3-5} \cmidrule(l){6-7}
\textbf{Task} & \textbf{Approach} & \textbf{Mass} & \textbf{Cent.} & \textbf{Cov.} & \textbf{FMD} & \textbf{1-NNA} \\
\midrule
\multirow{4}{0.8cm}{Right 
 Vent.} & Explicit & 154.5 & 227.1 & 101.4 & 164.7 & 0.761 \\
 & Implicit & 60.6 & 51.0 & 30.6 & 156.3 & 0.593 \\
 & Anatomica-L & 17.5 & 48.6 & 22.1 & 93.7 & \textbf{0.566} \\
 & Anatomica-V & \textbf{12.3} & \textbf{30.2} & \textbf{21.6} & \textbf{84.9} & 0.590 \\
\cline{1-7}
\multirow{4}{0.8cm}{Mitral 
Valve} & Explicit & 29.0 & 246.5 & 37.4 & 97.1 & 0.577 \\
 & Implicit & 8.91 & 87.0 & 17.3 & 314.8 & 0.661 \\
 & Anatomica-L & \textbf{3.22} & \textbf{11.4} & \textbf{7.89} & \textbf{88.8} & \textbf{0.577} \\
 & Anatomica-V & 3.81 & 41.4 & 7.99 & 93.8 & 0.586 \\
\cline{1-7}
\multirow{4}{0.8cm}{Aortic 
 Trunk} & Explicit & 2.53 & 272.6 & 13.9 & 114.3 & 0.587 \\
 & Implicit & \textbf{0.81} & \textbf{35.2} & \textbf{5.30} & 104.8 & 0.565 \\
 & Anatomica-L & 2.38 & 86.0 & 16.2 & 89.8 & 0.580 \\
 & Anatomica-V & 2.40 & 84.9 & 14.4 & \textbf{82.0} & \textbf{0.561} \\
\cline{1-7}
\multirow{4}{0.8cm}{Myo. 
 Wall} & Explicit & 1.01 & 123.4 & 3.39 & 130.7 & 0.574 \\
 & Implicit & 0.48 & \textbf{22.3} & \textbf{1.67} & 111.0 & 0.558 \\
 & Anatomica-L & \textbf{0.29} & 34.6 & 1.87 & \textbf{86.4} & 0.609 \\
 & Anatomica-V & 0.36 & 42.3 & 1.93 & 97.3 & \textbf{0.554} \\
\cline{1-7}
\end{tabular}
}
\end{table}

\subsection{Implementation Details}
As Anatomica guides the unconditional sampling process for each task in a training-free manner, we only train a separate unconditional diffusion model for each anatomical dataset. We present two variants of Anatomica that utilize different decoding and parsing strategies for guidance (\cref{tab:B0}). \textbf{Anatomica-V} uses a convolutional decoder to first produce a global voxel grid from the predicted clean latents, upon which we extract localized substructures during guidance with V-parsing. In contrast, \textbf{Anatomica-L} uses a neural field decoder to directly parse substructures with L-parsing. For the geometric control tasks, we use local L-parsing, while the topological tasks use coarse L-parsing. Unless specified otherwise, we use 100 diffusion sampling steps.

\subsection{Baselines}
 We also compare our general approach of guiding unconditional diffusion models against approaches that require conditional training for each task (\cref{tab:B0}). For the geometric control tasks, we implement conditional baselines to control target geometric features representing the size $m$, centroid $\mathbf{p}$, and covariance $\boldsymbol{\Sigma}$ of each substructure within the task. The first baseline is \textbf{Explicit Conditioning}, where we directly encode geometric attributes as scalar values in the conditioning signal \cite{kadry2024morphology,kadry2025diffusion}. We flatten and stack all geometric moments (mass, centroid, covariance) into a 13-dimensional vector for all $K$ substructures. We then expand this vector into a voxel grid $\mathbf{G}_\text{exp} \in \mathbb{R}^{13 \times K\times h \times w \times d}$ which is concatenated to the latent grid $\mathbf{z}$ along the channel dimension. The second is \textbf{Implicit Conditioning}, where we indirectly encode geometric attributes in the concatenated conditioning signal through 3D heatmaps \cite{kadry2025diffusion}. Here, we embed the geometric moments (centroid, covariance) as 3D Gaussians in voxel space. For each substructure, we create a voxel map $\mathbf{G}_\text{imp} \in \mathbb{R}^{K\times h \times w \times d}$ where the voxel values encode the Mahalanobis distance.
 \begin{figure}[h]
    \centering
    \includegraphics[width=\columnwidth]{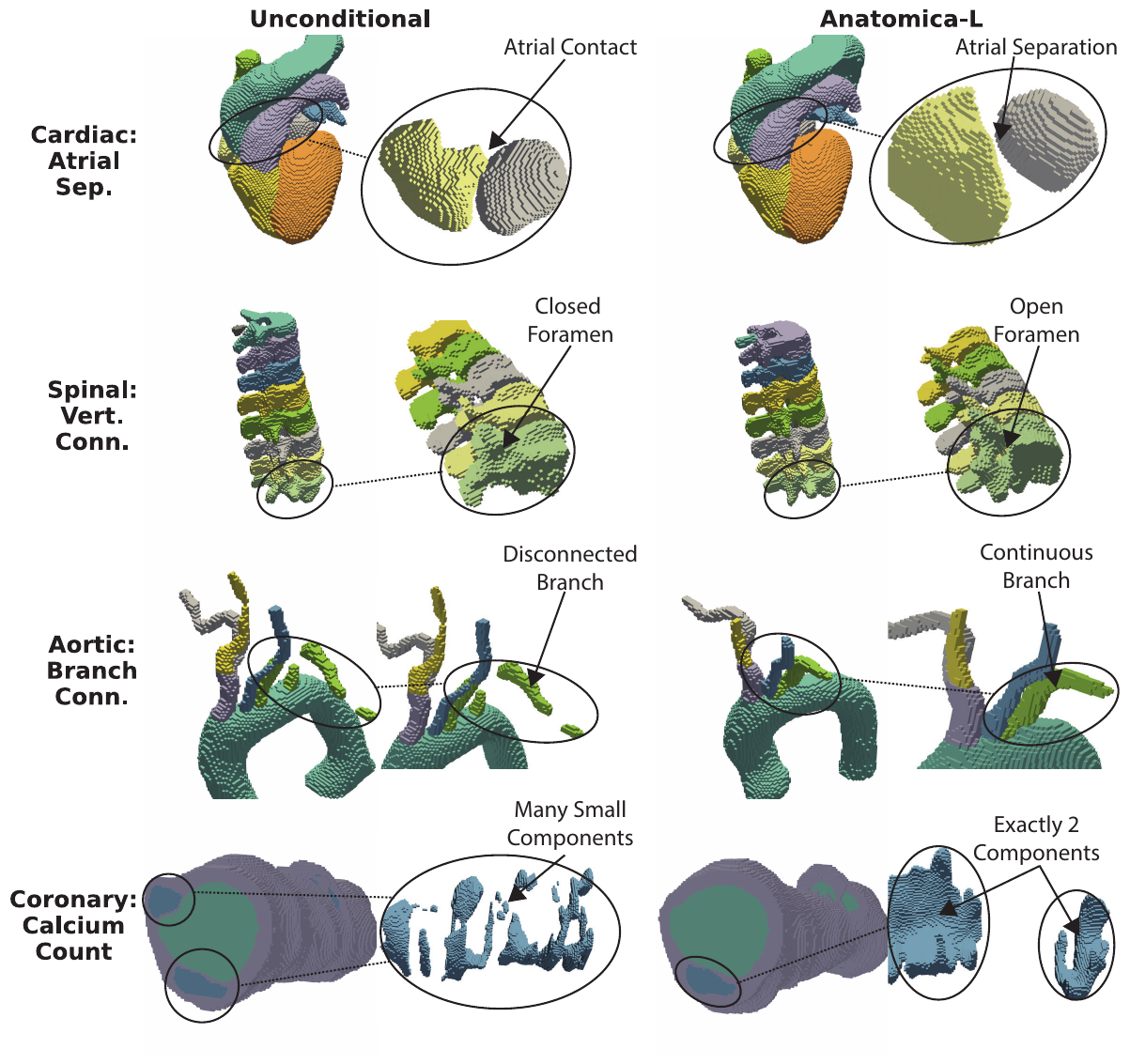}
    \caption{
        \textbf{Qualitative evaluation of topological control experiments.} We generate anatomical segmentations based on target topological priors for four anatomical datasets.
    }
    \label{fig:topo_control}
\end{figure}

 \renewcommand{\arraystretch}{1.25}
\begin{table}[h]
\centering
\caption{\textbf{Quantitative evaluation for topological control tasks}. We report Betti precision for number of connected components $B_0$, loops $B_1$, and voids $B_2$, and generation quality (1-NNA) for each approach.}
\label{tab:C1}
\resizebox{0.9\columnwidth}{!}{\begin{tabular}{cccccc}
\toprule
 &  & \multicolumn{3}{c}{\textbf{Topo. Precision (\%)}} & \textbf{Gen. Qual.} \\
 \cmidrule(r){3-5} \cmidrule(l){6-6}
\textbf{Task} & \textbf{Approach} & $\mathbf{B0}$ (↑) & $\mathbf{B1}$ (↑) & $\mathbf{B2}$ (↑) & \textbf{1-NNA (↓)} \\
\midrule
\multirow{2}{0.8cm}{Atrial 
 Sep.} & Uncond. & 7.81 & 5.47 & 56.2 & \textbf{0.578} \\
 & Anatomica-L & \textbf{78.9} & \textbf{89.1} & \textbf{97.7} & 0.606 \\
\cline{1-6}
\multirow{2}{0.8cm}{Branch 
 Conn.} & Uncond. & 55.5 & 12.5 & 63.3 & 0.559 \\
 & Anatomica-L & \textbf{77.3} & \textbf{17.2} & \textbf{64.1} & \textbf{0.532} \\
\cline{1-6}
\multirow{2}{0.8cm}{Vert. 
 Conn.} & Uncond. & 28.9 & 8.59 & \textbf{12.5} & \textbf{0.518} \\
 & Anatomica-L & \textbf{74.2} & \textbf{26.6} & 7.03 & 0.537 \\
\cline{1-6}
\multirow{2}{0.8cm}{Calcium 
 Count} & Uncond. & 0.00 & 2.34 & 95.3 & 0.653 \\
 & Anatomica-L & \textbf{60.9} & \textbf{79.7} & \textbf{98.4} & \textbf{0.618} \\
\cline{1-6}
\end{tabular}
}
\end{table}
 
\subsection{Evaluation Metrics}
We measure morphological quality metrics by measuring the Fréchet morphological distance (FMD) \cite{kadry2024probing} between real and synthetic distributions in morphological space. To compute such features, we consider all tissues as substructures and concatenate all masses, centroids, and normalized eigenvalues. We average the per-tissue 1-nearest neighbor accuracy (1-NNA) to compare point cloud distributions using the Earth Mover's Distance (EMD) \cite{yang2019pointflow}. We evaluate geometric control fidelity by taking the $L_1$-norm between the target and measured moments. Lastly, to measure topological control fidelity, we compute the topological precision for components $B_0$, loops $B_1$, and voids $B_2$ as the fraction of samples with the correct Betti number.
\subsection{Results}

\noindent\textbf{Precise Geometric Control} We quantitatively evaluate geometric control on the cardiac dataset with four different tasks. We see in \cref{tab:B1} and \cref{fig:geo_control} that our inference-time approach (Anatomica) is competitive in controlling the size, shape, position, and orientation of various substructures in the cardiac dataset. The closest baseline is the specialized implicit concatenation method, which requires conditional retraining for every task.
\begin{figure}[h]
    \centering
    \includegraphics[width=\columnwidth]{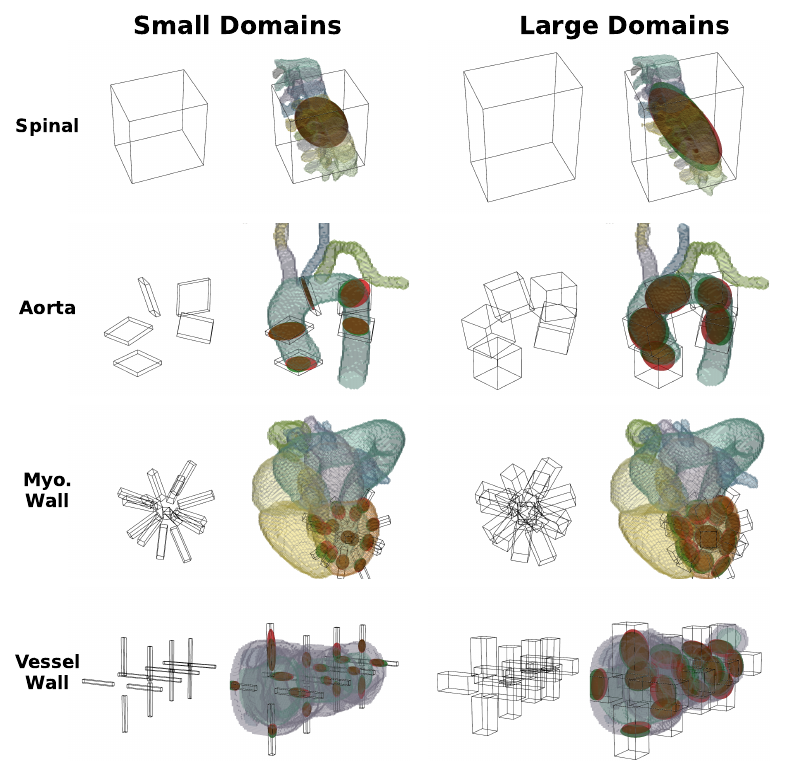}
    \caption{
        \textbf{Multi-scale geometric control of various anatomical substructures over different coordinate systems.} We generate anatomical segmentations based on domain size and anatomically relevant coordinate systems (Cartesian, curvilinear, cylindrical, and spherical).
    }
    \label{fig:flex_control}
\end{figure}

\noindent\textbf{Topological Control} We evaluate topological control on four anatomical datasets. We see in \cref{tab:C1} and \cref{fig:topo_control} that our framework is able to control the number of connected components and loops for various types of anatomical structures. Adherence to the number of voids is not improved for the aortic and vertebrae datasets, possibly due to the existence of single-voxel voids that are not easily detected at coarser measurement resolutions.

\noindent\textbf{Multi-scale Geometric Control} We demonstrate Anatomica's flexibility for geometric control across multiple anatomical types including the cardiac, aortic, spinal, and coronary datasets. As shown in \cref{fig:flex_control}, our framework handles varying spatial scales (large to small control domains) and operates across different coordinate system types (Cartesian, cylindrical, and spherical).

\noindent\textbf{Partial Decoding Ablation} We evaluate the speed-fidelity trade-off for partial decoding guidance under various decoding resolutions and strategies. We use the geometric control task involving the right ventricle for evaluation. We see in \cref{tab:D1} that low-resolution partial decoding with Anatomica-L maintains geometric fidelity while achieving substantial speedups against high-resolution decoding. We also find that given the same decoding resolution, using a neural field decoder (Anatomica-L) increases sampling speed at the cost of slightly reduced geometric fidelity, as compared to convolutional decoders (Anatomica-V).
\begin{table}[h]
    \centering
    \caption{\textbf{Quantitative ablation study for partial decoding strategies}. We evaluate the geometric fidelity and generation quality, and sampling speed for different decoding strategies and resolutions. Speed is measured in terms of sampled label maps per second using the maximum allowable batch size on a single GPU, normalized to the slowest method. Fidelity values for mass, centroid, and covariance are multiplied by 1e5, 1e4, 1e5 respectively.}
    \label{tab:D1}
    \resizebox{\columnwidth}{!}{\begin{tabular}{lllcccccc}
    \toprule
     \multicolumn{3}{c}{\textbf{Methodology}} & \multicolumn{3}{c}{\textbf{Geometric Fidelity (↓)}} & \multicolumn{2}{c}{\textbf{Gen. Qual. (↓)}} & \textbf{Speed (↑)} \\
     \cmidrule(r){1-3} \cmidrule(r){4-6} \cmidrule(l){7-9}
    \textbf{Approach} & \textbf{Domain} & \textbf{Res.} & \textbf{Mass} & \textbf{Cent.} & \textbf{Cov.} & \textbf{FMD} & \textbf{1-NNA} & \textbf{Speed} \\
    \midrule
    \multirow{6}{*}{Anatomica-L} & \multirow{3}{*}{Local} & High & 17.02 & \textbf{48.14} & \textbf{21.85} & \textbf{91.16} & 0.57 & 2.48 \\
     & & Med. & 16.64 & 48.41 & 22.03 & 93.89 & 0.58 & 7.43 \\
     & & Low & \textbf{16.43} & 48.75 & 22.07 & 105.57 & \textbf{0.55} & \textbf{10.40} \\
     \cmidrule(lr){2-9}
     & \multirow{3}{*}{Coarse} & High & \textbf{17.50} & 48.58 & \textbf{22.77} & \textbf{114.62} & 0.55 & 2.08 \\
     & & Med. & 17.75 & 48.78 & 23.03 & 119.93 & \textbf{0.53} & 7.43 \\
     & & Low & 20.30 & \textbf{46.96} & 25.60 & 123.31 & 0.54 & \textbf{10.40} \\
     \midrule
    Anatomica-V & Global & High & 11.95 & 30.66 & 21.92 & 84.70 & 0.58 & 1.00 \\
     \midrule
    \end{tabular}
    }
    \end{table}

\section{Discussion}
\noindent\textbf{Limitations} The main limitation of our guidance approach is that loss weightings should be tuned to balance the contributions of each constraint. However, we found that our choice of weights transfers readily between the four anatomical datasets we consider in this study. Moreover, we found that computing persistent homology at high resolution is computationally demanding and limits the coarse decoding resolution for topological guidance.

\noindent\textbf{Conclusion} We propose an inference-time framework for controlling generative models of anatomical voxel maps based on localized geo-topological attributes. Our design space centers on cuboidal control domains that compositionally slice out substructures across varying dimensions and coordinate systems. Over these domains, we propose the use of geo-topological potential functions for diffusion guidance, as well as neural field decoders for efficient partial decoding from latent space. We demonstrate state-of-the-art performance for geometric and topological control across a variety of anatomical systems and structures. We believe this work opens new avenues for controllable anatomical generation, with applications to virtual clinical trials and synthetic data augmentation for machine learning.


{
    \small
    \bibliographystyle{ieeenat_fullname}
    \bibliography{main}
}

\clearpage
\setcounter{page}{1}
\maketitlesupplementary
\section{Overview}
\noindent\textbf{Methods} In \cref{sec:supp_methods}, we detail the methods relating to diffusion model training, substructure parsing, and geo-topological measurement.

\noindent\textbf{Experimental Details} In \cref{sec:supp_exp_details}, we provide additional details on dataset creation, task setup, and evaluation metrics.

\noindent\textbf{Ablations} In \cref{sec:supp_ablations}, we study the influence of various hyperparamters such as individual loss weightings, decoding resolutions, and softmax temperature for geometric and topological guidance.
\section{Methodological Details}
\label{sec:supp_methods}
\subsection{Variational Autoencoder}
For this study, we adapt the voxel map VAE architecture specified by Kadry et al. \cite{kadry2025cardiocomposer}, which consists of a convolutional encoder and decoder. All architectural and training hyperparameters can be found in tables \ref{tab:VAE_architecture} and \ref{tab:VAE_training}.

\noindent\textbf{Decoder Architecture} We introduce two variants of Anatomica for latent diffusion guidance, with the primary difference being the decoder architecture that converts latent grid representation $\mathbf{z} \in \mathbb{R}^{c \times h \times w \times d}$ into a voxel grid representation $\hat{\mathbf{V}} \in \mathbb{R}^{C \times H \times W \times D}$ that can be anatomically characterized. \textbf{Anatomica-V} uses a convolutional decoder that mirrors the encoder, where the latent grid can only be decoded to full voxel resolution. On the other hand, \textbf{Anatomica-L} uses a neural field-based decoder that takes as input an arbitrary point grid $\mathbf{X}^q \in \mathbb{R}^{H \times W \times D \times 3}$ and returns for each point, the probability vector denoting the most likely anatomical class.

\noindent\textbf{Neural Field Decoder} Our decoder $\mathcal{F}$ decodes voxel maps with neural fields by first applying a bottleneck convolution to the latent grid representation in order to aggregate features within a local neighborhood. We then use a set of query points to interpolate into the latent grid representation using the slice operator $\mathcal{T}^l[\mathbf{X}^q]$ to create a set of latent points which are then point-wise concatenated with random Fourier features \cite{tancik2020fourier,long2021rffpytorch}. These features are then fed into a multi-layer perceptron (MLP) which consists of several hidden layers and finally outputs a logit vector for each query point. The logit vectors are then softmaxed to produce a segmentation probability vector.

\noindent\textbf{Training} We train all autoencoders with a combination of Dice-Cross Entropy reconstruction loss and KL divergence loss \cite{kadry2024morphology}. For neural field decoder training, we decode back to the full resolution global voxel grid with a query point grid $\mathbf{X}^q \in \mathbb{R}^{H \times W \times D \times 3}$.

\begin{table}[h]
    \centering
    \caption{Autoencoder architecture hyperparameters}
    \begin{tabular}{l|c}
    \hline
    \textbf{Conv. Encoder (shared)} & \textbf{Value} \\ \hline
    Num. Channels & [64, 128, 256] \\
    Num. Res. Blocks & 2 \\
    Final Downscaling Factor & 4 \\
    Bottleneck Dim & 3 \\
    \hline
    \textbf{Conv. Decoder (Anatomica-V)} & \textbf{Value}\\
    \hline
    Num. Channels & [64, 128, 256] \\
    Num. Res. Blocks & 2 \\
    Final Upscaling Factor & 4 \\
    \hline
    \textbf{Neural Field Decoder (Anatomica-L)} & \textbf{Value}\\
    \hline
    Bottleneck Conv. Channels & 64 \\
    Positional Encoding Dim & 10 \\
    Positional Encoding Bandwidth & 1 \\
    MLP Hidden Dim & 128 \\
    MLP Num Layers & 3 \\
    Normalization & LayerNorm \\
    Activation & ReLU \\
    \hline
    \end{tabular}
    \label{tab:VAE_architecture}
\end{table}

\begin{table}[h]
    \centering
    \caption{Autoencoder training hyperparameters}
    \begin{tabular}{l|c}
    \hline
    \textbf{Hyperparameter} & \textbf{Value} \\ \hline
    Learning Rate & $1 \times 10^{-5}$ \\
    Epochs & 40 \\
    Batch Size & 1 \\
    Dice-CE Loss Weight  & 1 \\
    KL Loss Weight & $1 \times 10^{-6}$ \\
    \hline
    \end{tabular}
    \label{tab:VAE_training}
\end{table}

\subsection{Latent Diffusion Model}
\noindent\textbf{Architecture \& Training}
For latent diffusion model architecture and training, we follow the formulation and architecture specified by Kadry et al. \cite{kadry2025cardiocomposer}.  All architectural and training hyperparameters can be found in table \ref{tab:LDM_config}. Our denoising model $D_\theta$ is parametrized in a skip-connection manner with a U-Net $\mathcal{K}_\theta$ with a convolutional encoder and decoder through the following relation:
\begin{equation}
\begin{split}
D_{\theta}(\mathbf{z}_\sigma;\sigma) &= c_{\text{skip}}(\sigma)\,\mathbf{z}_\sigma \\
&\quad + c_{\text{out}}(\sigma)\,\mathcal{K}_\theta\!\big(c_{\text{in}}(\sigma)\,\mathbf{z}_\sigma;\, c_{\text{noise}}(\sigma)\big)
\end{split}
\end{equation}
Where ($c_{\text{skip}}$,$c_{\text{out}}$,$c_{\text{in}}$,$c_{\text{noise}}$) are noise-level-dependent scaling coefficients \cite{karras2022elucidating}, $\sigma$ is the noise level. We use the same hyperparameters for the scaling coefficients as in \citet{karras2022elucidating}, but sample our noise level $p(\sigma)$ from a lognormal distribution with different parameters (see table \ref{tab:LDM_config}).

\noindent\textbf{Sampling}
Once the denoiser has been sufficiently trained, we define a specific noise level schedule governing the reverse process, in which the initial noise level, $\sigma$, starts at $\sigma_{\text{max}}$ and decreases to $\sigma_{\text{min}}$:
\begin{equation}
    \sigma_{i}=\left(\sigma_{\text{max}}^\frac{1}{\rho}+\frac{i}{N-1}(\sigma_{\text{min}}^\frac{1}{\rho}-\sigma_{\text{max}}^\frac{1}{\rho})\right)^\rho 
\end{equation}
where $\rho, \sigma_{min}$ and $\sigma_{max}$ are hyperparameters defined in table \ref{tab:LDM_config}. We specifically use the stochastic sampling method proposed in \citet{karras2022elucidating} (see \cref{tab:LDM_config} for hyperparameters).

\begin{table}[h]
    \centering
    \caption{Diffusion model hyperparameters}
    \begin{tabular}{c|c}
    \hline
    \textbf{Training}& \textbf{Value} \\ \hline
    lr & $2.5 \times 10^{-5}$ \\
    Epochs & 50 \\
    Batch Size & 1 \\
    Num. Channels & [64, 128, 196] \\
    Num. Res. Blocks & 2 \\
    Num. Attn. Heads & 1 \\
    Attn. Res. & 8, 4, 2 \\
    $\sigma_{\text{data}}$ & 1 \\
    $p(\sigma)$ mean & 1 \\
    $p(\sigma)$ std & 1.2 \\
    \hline
    \textbf{Sampling} & \textbf{Value} \\
    \hline
    $\sigma_{min}$ & $1 \times 10^{-2}$ \\
    $\sigma_{max}$ & 80 \\
    $\rho$ & 1 \\
    \hline
    \end{tabular}
    \label{tab:LDM_config}
\end{table}

\subsection{Substructure Parsing}
Anatomica revolves around parsing substructures through the use of selection vectors and control domains, enabling the measurement of anatomical properties for specified tissues within localized regions of interest. Selection vectors are binary vectors $\mathbf{u} \in \{0,1\}^C$ which select a subset of tissues from a voxel grid $\mathbf{V}$ through the Boolean subset operator $\mathcal{U}[\mathbf{u}]$ (see \cref{alg:boolean_subset}). By varying the Boolean selection vector, we enable the measurement of anatomic structures that are composed of multiple tissue types. Control domains are instantiated as point grids $\mathbf{X} \in \mathbb{R}^{\alpha \times \beta \times \gamma \times 3}$ that are used to parse substructures from grid-like representations such as voxel grids $\mathbf{V}$ using the voxel slicing operator $\mathcal{T}^s[\mathbf{X}]$ with \textbf{V-parsing} (see \cref{alg:voxel_parsing}). Alternatively, we can parse substructures directly from the latent representation $\mathbf{z}$ using the latent slicing operator $\mathcal{T}^l[\mathbf{X}]$ with \textbf{L-parsing} (see \cref{alg:latent_parsing}). To obtain control domains, we first define a template domain $\mathbf{X}^{\text{temp}} \in \mathbb{R}^{\alpha \times \beta \times \gamma \times 3}$ as a point grid centered at $\mathbf{0}$, with a grid size $\mathbf{g} = [\alpha, \beta, \gamma]$. We then apply a spatial transformation defined by affine transformation parameters $\mathbf{A}=[\mathbf{R}, \mathbf{s}, \mathbf{t}]$ to obtain anatomically relevant control domains.

\begin{algorithm}[H]
    \caption{Boolean Subset Operator}
    \label{alg:boolean_subset}
    \begin{algorithmic}[1]
    \small
    \Require $\mathbf{V} \in \mathbb{R}^{C \times H \times W \times D}$ \Comment{Voxel map}
    \Require $\mathbf{u} \in \{0,1\}^C$ \Comment{Boolean selection vector}
    \State $\hat{\mathbf{S}} \gets \mathbf{0}$ \Comment{Initialize}
    \For{tissue $i$ where $\mathbf{u}_i = 1$}
        \State $\hat{\mathbf{S}} \gets \max(\hat{\mathbf{S}}, \mathbf{V}_i)$ \Comment{Union via maximum}
    \EndFor
    \State \Return $\hat{\mathbf{S}} \in \mathbb{R}^{H \times W \times D}$
    \end{algorithmic}
    \end{algorithm}

\begin{algorithm}[H]
\caption{Voxel Substructure Parsing (V-parsing)}
\label{alg:voxel_parsing}
\begin{algorithmic}[1]
\small
\Require $\mathbf{V} \in \mathbb{R}^{C \times H \times W \times D}$ \Comment{Voxel map}
\Require $\mathbf{u} \in \{0,1\}^C$ \Comment{Selection vector}
\Require $\{\mathbf{X}_k\}_{k=1}^K$ where $\mathbf{X}_k \in \mathbb{R}^{\alpha \times \beta \times \gamma \times 3}$ \Comment{Control domains}
\Algphase{Subset Tissues}
\State $\hat{\mathbf{S}} \gets \mathcal{U}[\mathbf{u}](\mathbf{V}) \in \mathbb{R}^{H \times W \times D}$ \Comment{Boolean subset}
\Algphase{Parse Substructures}
\For{$k = 1, \ldots, K$}
    \State $\mathbf{S}_k \gets \mathcal{T}^s[\mathbf{X}_k](\hat{\mathbf{S}}) \in \mathbb{R}^{\alpha \times \beta \times \gamma}$ \Comment{Voxel slice}
\EndFor
\State \Return $\{\mathbf{S}_k\}_{k=1}^K$
\end{algorithmic}
\end{algorithm}

\begin{algorithm}[H]
\caption{Latent Substructure Parsing (L-parsing)}
\label{alg:latent_parsing}
\begin{algorithmic}[1]
\small
\Require $\mathbf{z} \in \mathbb{R}^{c \times h \times w \times d}$ \Comment{Latent representation}
\Require $\mathbf{u} \in \{0,1\}^C$ \Comment{Selection vector}
\Require $\{\mathbf{X}_k\}_{k=1}^K$ where $\mathbf{X}_k \in \mathbb{R}^{\alpha \times \beta \times \gamma \times 3}$ \Comment{Control domains}
\Algphase{Parse Substructures}
\For{$k = 1, \ldots, K$}
    \State $\mathbf{z}_k \gets \mathcal{T}^l[\mathbf{X}_k](\mathbf{z}) \in \mathbb{R}^{c \times \alpha \times \beta \times \gamma}$ \Comment{Latent slice}
    \State $\mathbf{S}_k \gets \mathcal{U}[\mathbf{u}] \circ \mathcal{F}[\mathbf{X}_k](\mathbf{z}_k) \in \mathbb{R}^{\alpha \times \beta \times \gamma}$ \Comment{Decode \& subset}
\EndFor
\State \Return $\{\mathbf{S}_k\}_{k=1}^K$
\end{algorithmic}
\end{algorithm}

\subsection{Control Domains}
\label{sec:control_domain_algorithms}
Anatomica supports several methods for defining control domains $\mathbf{X}_k$, each useful for probing different anatomical or geometric properties. In this study, we primarily compute control domain parameters from real anatomical voxel maps and measure geometric properties within such domains to define targets for diffusion guidance. This approach is not limited to guidance use-cases, and can potentially be used for other machine-learning tasks that use differentiable loss functions. We now detail the algorithmic procedures for computing control domain parameters across different coordinate systems from anatomical voxel maps. 


\noindent\textbf{Global Domain Computation}
The global control domain can be used to measure properties over the entire voxel grid without geometric feature extraction at a variable spatial resolution. We compute global domains through \cref{alg:global}.

\begin{algorithm}[H]
\caption{Global Domain Computation}
\label{alg:global}
\begin{algorithmic}[1]
\small
\Require $(\alpha, \beta, \gamma)$ with $\alpha \approx \beta \approx \gamma$ \Comment{Volumetric grid size}
\Algphase{Set Affine Parameters}
\State $\mathbf{R} \gets \mathbf{I} \in \mathbb{R}^{3 \times 3}$ \Comment{Rotation}
\State $\mathbf{t} \gets \mathbf{0} \in \mathbb{R}^{3}$ \Comment{Translation}
\State $\mathbf{s} \gets \mathbf{1} \in \mathbb{R}^{3}$ \Comment{Scale}
\State \Return $\mathbf{A} = [\mathbf{R}, \mathbf{s}, \mathbf{t}]$
\end{algorithmic}
\end{algorithm}

\noindent\textbf{Cartesian Domain Computation}
Cartesian domains enable the measurement of anatomical properties within localized bounding boxes that contain structures of interest. We compute Cartesian domains through \cref{alg:cartesian}.

\begin{algorithm}[H]
\caption{Cartesian Domain Computation}
\label{alg:cartesian}
\begin{algorithmic}[1]
\small
\Require $\mathbf{u} \in \{0,1\}^C$ \Comment{Tissue selection vector}
\Require $(\alpha, \beta, \gamma)$ with $\alpha \approx \beta \approx \gamma$ \Comment{Volumetric grid size}
\Algphase{Extract Bounding Box}
\State $\hat{\mathbf{S}} \gets \mathcal{U}[\mathbf{u}](\mathbf{V})$, $\tilde{\mathbf{S}} \gets \mathbb{I}[\hat{\mathbf{S}} > 0.9]$ \Comment{Subset \& binarize}
\State $\mathbf{r}^{\text{upper}}, \mathbf{r}^{\text{lower}} \gets \text{ExtractLimits}(\tilde{\mathbf{S}})$ where $\mathbf{r}^{\text{upper}}, \mathbf{r}^{\text{lower}} \in \mathbb{R}^3$
\Algphase{Set Affine Parameters}
\State $\mathbf{R} \gets \mathbf{I} \in \mathbb{R}^{3 \times 3}$ \Comment{Rotation}
\State $\mathbf{t} \gets (\mathbf{r}^{\text{upper}} + \mathbf{r}^{\text{lower}})/2 \in \mathbb{R}^{3}$ \Comment{Translation}
\State $\mathbf{s} \gets (\mathbf{r}^{\text{upper}} - \mathbf{r}^{\text{lower}}) \oslash [\alpha, \beta, \gamma]^T \in \mathbb{R}^{3}$ \Comment{Scale}
\State \Return $\mathbf{A} = [\mathbf{R}, \mathbf{s}, \mathbf{t}]$
\end{algorithmic}
\end{algorithm}

\noindent\textbf{Interface Domain Computation}
Interface domains enable the measurement of local anatomical properties at the interfacial region between two or more structures, such as valve annuli or branch points. We compute interface domains through \cref{alg:interface}.

\begin{algorithm}[H]
\caption{Interface Domain Computation}
\label{alg:interface}
\begin{algorithmic}[1]
\small
\Require $\mathbf{u}^A, \mathbf{u}^B \in \{0,1\}^C$ \Comment{Tissue selection vectors}
\Require $(\alpha, \beta, \gamma)$ with $\alpha \ll \beta \approx \gamma$ \Comment{Planar grid size}
\Require $k_{\text{dil}}$, $\mathbf{R}^r \in \mathbb{R}^{3 \times 1}$ \Comment{Kernel size, ref vector}
\Algphase{Extract Interface Regions}
\State $\hat{\mathbf{S}}^A \gets \mathcal{U}[\mathbf{u}^A](\mathbf{V})$, $\hat{\mathbf{S}}^B \gets \mathcal{U}[\mathbf{u}^B](\mathbf{V})$ \Comment{Subset}
\State $\hat{\mathbf{S}}^A_{\text{dil}} \gets \text{maxpool}_{k_{\text{dil}}}(\hat{\mathbf{S}}^A)$, $\hat{\mathbf{S}}^B_{\text{dil}} \gets \text{maxpool}_{k_{\text{dil}}}(\hat{\mathbf{S}}^B)$ \Comment{Dilate}
\State $\mathbf{M} \gets \min(\hat{\mathbf{S}}^A_{\text{dil}}, \hat{\mathbf{S}}^B_{\text{dil}})$ \Comment{Combine}
\State $\hat{\mathbf{S}}^A_{\text{int}} \gets \hat{\mathbf{S}}^A \odot \mathbf{M}$, $\hat{\mathbf{S}}^B_{\text{int}} \gets \hat{\mathbf{S}}^B \odot \mathbf{M}$ \Comment{Mask interface}
\Algphase{Compute Interface Frame Orientations}
\State $\mathbf{p}^A, \mathbf{p}^B \gets \text{Centroid}(\hat{\mathbf{S}}^A_{\text{int}}), \text{Centroid}(\hat{\mathbf{S}}^B_{\text{int}})$ \Comment{(Alg. \ref{alg:geometric_measurement})}
\State $\mathbf{R}^\alpha \gets (\mathbf{p}^B - \mathbf{p}^A) / \|\mathbf{p}^B - \mathbf{p}^A\| \in \mathbb{R}^{3 \times 1}$ \Comment{Interface vector}
\State $\mathbf{R}^\beta, \mathbf{R}^\gamma \gets \text{Orthonorm.}(\mathbf{R}^\alpha, \mathbf{R}^r) \in \mathbb{R}^{3 \times 1}$ \Comment{(Alg. \ref{alg:orthonormalize})}
\Algphase{Set Affine Parameters}
\State $\mathbf{R} \gets [\mathbf{R}^\alpha, \mathbf{R}^\beta, \mathbf{R}^\gamma] \in \mathbb{R}^{3 \times 3}$ \Comment{Rotation}
\State $\mathbf{s} \gets [\alpha/H, \beta/W, \gamma/D]^T \in \mathbb{R}^{3}$ \Comment{Scale}
\State $\mathbf{t}^A \gets \mathbf{p}^A \in \mathbb{R}^{3}$, $\mathbf{t}^B \gets \mathbf{p}^B \in \mathbb{R}^{3 \times 1}$ \Comment{Translation}
\State \Return $\mathbf{A}^A = [\mathbf{R}, \mathbf{s}, \mathbf{t}^A]$, $\mathbf{A}^B = [\mathbf{R}, \mathbf{s}, \mathbf{t}^B]$
\end{algorithmic}
\end{algorithm}

\noindent\textbf{Curvilinear Domain Computation}
Curvilinear domains enable the measurement of cross-sectional anatomical properties along tubular structures such as blood vessels. We compute curvilinear domains through \cref{alg:curvilinear}. For skeletonization, we follow the methods and hyperparameters detailed in \citet{kadry2024morphology} for non-differentiable hard skeletonization. 

\begin{algorithm}[H]
\caption{Curvilinear Domain Computation}
\label{alg:curvilinear}
\begin{algorithmic}[1]
\small
\Require $\mathbf{u} \in \{0,1\}^C$ \Comment{Tissue selection vector}
\Require $(\alpha, \beta, \gamma)$ with $\alpha \ll \beta \approx \gamma$ \Comment{Planar grid size}
\Require $\mathbf{i}_{\text{sub}}$, $\mathbf{R}^r \in \mathbb{R}^{3 \times 1}$ \Comment{Subsampling Indices, Ref vector}
\Algphase{Extract Centerline}
\State $\hat{\mathbf{S}} \gets \mathcal{U}[\mathbf{u}](\mathbf{V})$, $\tilde{\mathbf{S}} \gets \mathbb{I}[\hat{\mathbf{S}} > 0.9]$ \Comment{Subset \& binarize}
\State $\mathbf{C} \gets \text{Skeletonize}(\tilde{\mathbf{S}})$ where $\mathbf{C} \in \mathbb{R}^{N_{\text{center}} \times 3}$
\Algphase{Compute Curvilinear Frames}
\State $\mathbf{F}^\alpha \gets \text{FiniteDifference}(\mathbf{C}) \in \mathbb{R}^{N_{\text{center}} \times 3}$ \Comment{Tangent vectors}
\State $\mathbf{F}^\beta_0, \mathbf{F}^\gamma_0 \gets \text{Orthonorm.}(\mathbf{F}^\alpha_0, \mathbf{R}^r)$ \Comment{(Alg. \ref{alg:orthonormalize})}
\State $\mathbf{F}^\beta, \mathbf{F}^\gamma \gets \text{ParallelTransport}(\mathbf{F}^\alpha, \mathbf{F}^\beta_0, \mathbf{F}^\gamma_0)$ \Comment{(Alg. \ref{alg:parallel_transport})}
\State $\mathbf{C}^{\text{sub}} \gets \text{Subsample}(\mathbf{C}, \mathbf{i}_{\text{sub}})$ where $\mathbf{C}^{\text{sub}} \in \mathbb{R}^{N_{\text{planes}} \times 3}$
\State $\mathbf{R}^\alpha, \mathbf{R}^\beta, \mathbf{R}^\gamma \gets \text{Subsample}(\mathbf{F}^\alpha, \mathbf{F}^\beta, \mathbf{F}^\gamma, \mathbf{i}_{\text{sub}}) \in \mathbb{R}^{N_{\text{planes}} \times 3}$
\Algphase{Set Affine Parameters}
\For{domain $k = 1, \ldots, N_{\text{planes}}$}
    \State $\mathbf{R}_k \gets [\mathbf{R}^\alpha_k, \mathbf{R}^\beta_k, \mathbf{R}^\gamma_k] \in \mathbb{R}^{3 \times 3}$ \Comment{Rotation}
    \State $\mathbf{s}_k \gets [\alpha/H, \beta/W, \gamma/D]^T \in \mathbb{R}^{3}$ \Comment{Scale}
    \State $\mathbf{t}_k \gets \mathbf{C}^{\text{sub}}_k \in \mathbb{R}^{3}$ \Comment{Translation}
\EndFor
\State \Return $\{\mathbf{A}_k\}_{k=1}^{N_{\text{planes}}}$
\end{algorithmic}
\end{algorithm}

\noindent\textbf{Spherical Domain Computation}
Spherical domains enable the measurement of radial anatomical properties of shell-like structures such as myocardial walls. We compute spherical domains through \cref{alg:spherical}. Instead of sampling equidistant points in polar and azimuthal space, we compute equally distributed points on the sphere surface using the Fibonacci lattice algorithm \cite{gonzalez2010measurementfibonacci}.

\begin{algorithm}[H]
\caption{Spherical Domain Computation}
\label{alg:spherical}
\begin{algorithmic}[1]
\small
\Require $\mathbf{u} \in \{0,1\}^C$ \Comment{Tissue selection vector}
\Require $(\alpha, \beta, \gamma)$ with $\alpha \approx \beta \ll \gamma$ \Comment{Ray-like grid size}
\Require $N_{\text{rays}}$, $N_q$, $\mathbf{R}^r \in \mathbb{R}^{3 \times 1}$ \Comment{Number of rays, query ray resolution, ref vector}
\Algphase{Generate Radial Directions}
\State $\hat{\mathbf{S}} \gets \mathcal{U}[\mathbf{u}](\mathbf{V})$ \Comment{Subset tissues}
\State $\mathbf{p} \gets \text{Centroid}(\hat{\mathbf{S}}) \in \mathbb{R}^{3}$ \Comment{(Alg. \ref{alg:geometric_measurement})}
\State $\mathbf{R}^\gamma \gets \text{FibonacciLattice}(N_{\text{rays}}, \mathbf{p})$ where $\mathbf{R}^\gamma \in \mathbb{R}^{N_{\text{rays}} \times 3}$
\State $\mathbf{R}^\beta, \mathbf{R}^\alpha \gets \text{Orthonorm.}(\mathbf{R}^\gamma,\mathbf{R}^r) \in \mathbb{R}^{N_{\text{rays}} \times 3}$\Comment{(Alg. \ref{alg:orthonormalize})}
\Algphase{Find Wall Centroids and Set Affine Parameters}
\For{domain $k = 1, \ldots, N_{\text{rays}}$}
    \State $\mathbf{X}^{\text{ray}}_k \gets \text{MakeQueryRay}(\mathbf{p}, \mathbf{R}^\gamma_k, N_q)$ where $\mathbf{X}^{\text{ray}}_k \in \mathbb{R}^{1 \times 1 \times N_q \times 3}$
    \State $\mathbf{S}^{\text{ray}}_k \gets \mathcal{T}^s[\mathbf{X}^{\text{ray}}_k](\hat{\mathbf{S}})$ \Comment{Slice along ray}
    \State $\mathbf{p}_{\text{wall},k} \gets \text{Centroid}(\mathbf{S}^{\text{ray}}_k) \in \mathbb{R}^{3}$ \Comment{(Alg. \ref{alg:geometric_measurement})}
    \State $\mathbf{R}_k \gets [\mathbf{R}^\alpha_k, \mathbf{R}^\beta_k, \mathbf{R}^\gamma_k] \in \mathbb{R}^{3 \times 3}$ \Comment{Rotation}
    \State $\mathbf{s}_k \gets [\alpha/H, \beta/W, \gamma/D]^T \in \mathbb{R}^{3}$ \Comment{Scale}
    \State $\mathbf{t}_k \gets \mathbf{p}_{\text{wall},k} \in \mathbb{R}^{3}$ \Comment{Translation}
\EndFor
\State \Return $\{\mathbf{A}_k\}_{k=1}^{N_{\text{rays}}}$
\end{algorithmic}
\end{algorithm}

\noindent\textbf{Cylindrical Domain Computation}
Cylindrical domains enable the measurement of radial anatomical properties of walled tubular structures such as coronary arteries. We compute cylindrical domains through \cref{alg:cylindrical}. We compute cylindrical domains by defining equidistant ray centers along the z-axis and equally sampling the polar directions according to predefined sampling resolutions.

\begin{algorithm}[H]
\caption{Cylindrical Domain Computation}
\label{alg:cylindrical}
\begin{algorithmic}[1]
\small
\Require $\mathbf{u} \in \{0,1\}^C$ \Comment{Tissue selection vector}
\Require $(\alpha, \beta, \gamma)$ with $\alpha \approx \beta \ll \gamma$ \Comment{Ray-like grid size}
\Require $N_z$, $N_\theta$, $N_q$, $\mathbf{R}^r \in \mathbb{R}^{3 \times 1}$ \Comment{Z-levels, angles, query ray resolution, ref vector}
\Algphase{Generate Cylindrical Directions}
\State $\hat{\mathbf{S}} \gets \mathcal{U}[\mathbf{u}](\mathbf{V})$ \Comment{Subset tissues}
\State $\mathbf{R}^\gamma \gets \text{CylindricalLattice}(N_z, N_\theta)$ where $\mathbf{R}^\gamma \in \mathbb{R}^{N_{\text{rays}} \times 3}$, $N_{\text{rays}} = N_z \times N_\theta$
\State $\mathbf{R}^\beta, \mathbf{R}^\alpha \gets \text{Orthonorm.}(\mathbf{R}^\gamma, \mathbf{R}^r) \in \mathbb{R}^{N_{\text{rays}} \times 3}$ \Comment{(Alg. \ref{alg:orthonormalize})}
\Algphase{Find Wall Centroids and Set Affine Parameters}
\For{domain $k = 1, \ldots, N_{\text{rays}}$}
    \State $\mathbf{X}^{\text{ray}}_k \gets \text{MakeQueryRay}(\mathbf{R}^\gamma_k, N_q)$ where $\mathbf{X}^{\text{ray}}_k \in \mathbb{R}^{1 \times 1 \times N_q \times 3}$
    \State $\mathbf{S}^{\text{ray}}_k \gets \mathcal{T}^s[\mathbf{X}^{\text{ray}}_k](\hat{\mathbf{S}})$ \Comment{Slice along ray}
    \State $\mathbf{p}_{\text{wall},k} \gets \text{Centroid}(\mathbf{S}^{\text{ray}}_k) \in \mathbb{R}^{3}$ \Comment{(Alg. \ref{alg:geometric_measurement})}
    \State $\mathbf{R}_k \gets [\mathbf{R}^\alpha_k, \mathbf{R}^\beta_k, \mathbf{R}^\gamma_k] \in \mathbb{R}^{3 \times 3}$ \Comment{Rotation}
    \State $\mathbf{s}_k \gets [\alpha/H, \beta/W, \gamma/D]^T \in \mathbb{R}^{3}$ \Comment{Scale}
    \State $\mathbf{t}_k \gets \mathbf{p}_{\text{wall},k} \in \mathbb{R}^{3}$ \Comment{Translation}
\EndFor
\State \Return $\{\mathbf{A}_k\}_{k=1}^{N_{\text{rays}}}$
\end{algorithmic}
\end{algorithm}

\noindent\textbf{Parallel Transport Procedure}
For curvilinear coordinate systems, we aim to maintain consistent frame orientations as we move along the centerline. To do this, we apply parallel transport by propagating an initial orthonormal frame along a centerline using the Rodrigues rotation formula.

\begin{algorithm}[H]
\caption{ParallelTransport}
\label{alg:parallel_transport}
\begin{algorithmic}[1]
\small
\Require $\mathbf{F}^1 \in \mathbb{R}^{N_{\text{center}} \times 3}$ \Comment{Normalized tangent vectors}
\Require $\mathbf{F}^2_0, \mathbf{F}^3_0 \in \mathbb{R}^{3}$ \Comment{Initial normalized frame vectors}
\For{$i = 1, \ldots, N_{\text{center}} - 1$}
    \State $\mathbf{a}_i \gets (\mathbf{F}^1_{i-1} \times \mathbf{F}^1_i) / \|\mathbf{F}^1_{i-1} \times \mathbf{F}^1_i\|$ \Comment{Rotation axis}
    \State $\theta_i \gets \cos^{-1}(\mathbf{F}^1_{i-1} \cdot \mathbf{F}^1_i)$ \Comment{Rotation angle}
    \State $\mathbf{F}^2_i, \mathbf{F}^3_i \gets \text{Rodrigues}(\mathbf{F}^2_{i-1}, \mathbf{F}^3_{i-1}, \mathbf{a}_i, \theta_i)$
\EndFor
\State \Return $\mathbf{F}^2, \mathbf{F}^3 \in \mathbb{R}^{N_{\text{center}} \times 3}$
\end{algorithmic}
\end{algorithm}

\noindent\textbf{Orthonormalization Procedure}
For interface, curvilinear, spherical, and cylindrical coordinate systems, we wish to compute a set of orthonormal frame vectors from an initial vector. To do this, we define an arbitrary reference vector $\mathbf{R}^r$ and compute orthonormal frame vectors from a primary direction vector by taking successive cross products. For numerical stability, we use an alternate reference vector if the reference and initial vectors are perfectly aligned.

\begin{algorithm}[H]
\caption{Orthonormalization}
\label{alg:orthonormalize}
\begin{algorithmic}[1]
\small
\Require $\mathbf{U}^0 \in \mathbb{R}^{3 \times 1}$ \Comment{Primary direction vector}
\Require $\mathbf{R}^r \in \mathbb{R}^{3 \times 1}$ \Comment{Reference vector}
\State $\mathbf{U}^1 \gets (\mathbf{U}^0 \times \mathbf{R}^r) / \|\mathbf{U}^0 \times \mathbf{R}^r\| \in \mathbb{R}^{3 \times 1}$ \Comment{Second frame vector}
\State $\mathbf{U}^2 \gets \mathbf{U}^0 \times \mathbf{U}^1 \in \mathbb{R}^{3 \times 1}$ \Comment{Third frame vector (auto-normalized)}
\State \Return $\mathbf{U}^1, \mathbf{U}^2$
\end{algorithmic}
\end{algorithm}

\subsection{Geometric Measurement \& Guidance}
\noindent\textbf{Scale Standardization of Mass}
We normalize the measured mass $m_k$ by the total number of voxels in the control domain $\alpha \beta \gamma$ in order to remain invariant to control domain discretization, allowing us to maintain similar geometric loss weightings across different discretization levels.

\noindent\textbf{Local to Global Transformation of Moments}
Our geometric moment formulation can be sensitive to control domain discretization and coordinate system choice. For example, our substructure can inhabit 80\% of the control domain, but the control domain may be a small region within the global domain, resulting in a large measured mass $m_k$. Another example would be the case of a localized control domain with a measured centroid $\mathbf{p}_k$ that is measured to be in the center of the control domain, but is at the periphery of the global domain. We therefore aim to express our geometric measurements in a manner that is invariant to control domain choice. This is important when applying MSE-based geometric loss functions across different tasks due to varying scales. Therefore, we express all geometric moments in the global coordinate system using the inverse of the control domain transformation parameters $\mathbf{A}_k=[\mathbf{R}_k, \mathbf{s}_k, \mathbf{t}_k]$.

\noindent\textbf{Stabilizing Covariance Normalization} As we normalize the covariance matrix by the trace, we stabilize the gradient in the case of empty substructures by adding a small epsilon (1e-9) to the diagonal of the covariance matrix.

\noindent\textbf{Adaptive Mass Weighting} To avoid centroid and covariance gradient explosion in the case of near-empty segmentations, we adaptively weight the centroid and covariance losses by the mass of the substructure. Below a predefined threshold, we set the centroid and covariance weightings $\lambda_1 = \lambda_2 = 0$. This mass threshold is determined on a task-by-task basis, where we multiply the average measured mass for the task by a factor of 0.1. 

\begin{algorithm}[H]
    \caption{Geometric Measurement}
    \label{alg:geometric_measurement}
    \begin{algorithmic}[1]
    \small
    \Require $\mathbf{S}_k \in \mathbb{R}^{\alpha \times \beta \times \gamma}$ \Comment{Substructure}
    \Require $\mathbf{R}_k \in \mathbb{R}^{3 \times 3}$ \Comment{Rotation matrix}
    \Require $\mathbf{s}_k \in \mathbb{R}^3$ \Comment{Scale vector}
    \Require $\mathbf{t}_k \in \mathbb{R}^3$ \Comment{Translation vector}
    \Algphase{Compute Local Moments}
    \State $m_k^{\text{raw}} \gets \text{ComputeMass}(\mathbf{S}_k)$ \Comment{(Eq. \ref{eq:geometric_moments})}
    \State $\mathbf{p}_k^{\text{local}} \gets \text{ComputeCentroid}(\mathbf{S}_k, m_k^{\text{raw}})$ \Comment{(Eq. \ref{eq:geometric_moments})}
    \State $\boldsymbol{\Sigma}_k^{\text{local}} \gets \text{ComputeCovariance}(\mathbf{S}_k, \mathbf{p}_k^{\text{local}}, m_k^{\text{raw}})$ \Comment{(Eq. \ref{eq:geometric_moments})}
    \State $m_k^{\text{local}} \gets m_k^{\text{raw}} / (\alpha\beta\gamma)$ \Comment{Normalize by voxel count}
    \Algphase{Local to Global Transformation}
    \State $\mathbf{J}_k \gets \mathbf{R}_k \operatorname{diag}(\mathbf{s}_k)$ \Comment{Rotation-scale matrix}
    \State $m_k^{\text{global}} \gets m_k^{\text{local}} \cdot |\det(\mathbf{J}_k)|$ \Comment{Transform mass}
    \State $\mathbf{d}_k^{\text{local}} \gets \mathbf{p}_k^{\text{local}} - \frac{1}{2}\mathbf{1}$ \Comment{Local displacement from center}
    \State $\mathbf{d}_k^{\text{global}} \gets \mathbf{J}_k \mathbf{d}_k^{\text{local}}$ \Comment{Transform displacement}
    \State $\mathbf{p}_k^{\text{global}} \gets \mathbf{t}_k + \mathbf{d}_k^{\text{global}}$ \Comment{Transform centroid}
    \State $\boldsymbol{\Sigma}_k^{\text{global}} \gets \mathbf{J}_k \boldsymbol{\Sigma}_k^{\text{local}} \mathbf{J}_k^T$ \Comment{Transform covariance}
    \State \Return $(m_k^{\text{global}}, \mathbf{p}_k^{\text{global}}, \boldsymbol{\Sigma}_k^{\text{global}})$
    \end{algorithmic}
    \end{algorithm}
    
\subsection{Topological Measurement \& Guidance}
We partition the persistence set into disjoint sets $\mathcal{Y}_k$ and $\mathcal{Z}_k$ consisting of points that should be preserved or suppressed based on a topological prior $\mathcal{B}_k \in \mathbb{R}^{3} =[\mathbf{B}0,\mathbf{B}1,\mathbf{B}2]$, which specifies the desired features for the components, loops, and voids, respectively. For each dimension, we sort the points by persistence and select the top $\mathcal{B}_i$ points for each dimension $i$ specified by the prior. 
\subsection{Parallelization}
For our geometric measurement operations, we take advantage of parallel GPU computation. We parallelize across different batch indices, constraints, and substructures. When computing control domains, some domain types allow for invalid domains, such as in the case of spherical ray domains, where the ray may not intersect with the substructure. We handle these invalid domains by masking out the computed loss. The only exception is our skeletonization step for curvilinear control domains, as our implementation is computed on a CPU. For topological measurement, we do not parallelize the persistent homology computation as no GPU-compatible implementation is publicly available, and CPU-parallelization over several cores did not provide significant speedups.
\section{Experimental Details}
\label{sec:supp_exp_details}
\subsection{Baselines}
\noindent\textbf{Explicit Conditioning} To ensure that the elements of $\mathcal{G}\text{exp}$ are roughly between 0 and 1, we min-max normalize the masses $m_k$, centroids $\mathbf{p}_k$, and normalized covariances $\boldsymbol{\Sigma}_k^n$ with values calculated from the real dataset (\cref{tab:geom_norm_values}). The LDM input channel count is increased to accommodate the concatenated input. This method does not readily permit the use of dropout to train a diffusion model in an unconditional manner because the null condition is defined as zero, which is equivalent to the minimum moment values.

\noindent\textbf{Implicit Conditioning} To compute the ellipsoidal distance map, we use the centroids $\mathbf{p}_k$ and non-normalized covariances $\boldsymbol{\Sigma}_k$ for each component to compute the Mahalanobis distance \cite{de2000mahalanobis} for each voxel position. We then apply a shifted sigmoid transform to constrain the outputs between 0 and 1, and subsequently concatenate the resulting grid to the latents. To enable unconditional generation, we randomly drop out each substructure channel of $\mathcal{G}\text{imp}$ with a probability of 0.1.

\begin{table}[t]
    \centering
    \caption{Normalizing constants for geometric moments during explicit conditioning across different tasks.}
    \label{tab:geom_norm_values}
    \resizebox{\columnwidth}{!}{
    \begin{tabular}{l|cccc}
    \toprule
    \multicolumn{1}{c|}{} & \multicolumn{4}{c}{\textbf{Geometric Control Task}} \\
    \hline
    \textbf{Parameter} & \textbf{RV} & \textbf{Mitral} & \textbf{Aortic} & \textbf{Myo} \\
    \midrule
    Mass Min $m_k$ & $3.19 \times 10^{-3}$ & $3.67 \times 10^{-4}$ & $0$ & $0$ \\
    Mass Max $m_k$ & $1.3 \times 10^{-2}$ & $1.36 \times 10^{-3}$ & $8.59 \times 10^{-4}$ & $1.95 \times 10^{-5}$ \\
    \midrule
    Centroid Min $\mathbf{p}_k$ & $0$ & $0$ & $-7.81 \times 10^{-3}$ & $0$ \\
    Centroid Max $\mathbf{p}_k$ & $1$ & $1$ & $0.91$ & $0.64$ \\
    \midrule
    Covariance Min $\boldsymbol{\Sigma}_k$ & $-1 \times 10^{-4}$ & $-8.66 \times 10^{-4}$ & $-5.59 \times 10^{-4}$ & $2.88 \times 10^{-4}$ \\
    Covariance Max $\boldsymbol{\Sigma}_k$ & $1 \times 10^{-2}$ & $2.34 \times 10^{-3}$ & $1.56 \times 10^{-3}$ & $8.03 \times 10^{-4}$ \\
    \bottomrule
    \end{tabular}
    }
\end{table}
\subsection{Datasets}
\noindent\textbf{Cardiac Dataset} For our study, we utilize TotalSegmentator v2 ~\cite{wasserthal2022totalsegmentator} to create the cardiac segmentations, with 596 3D segmentations manually selected based on segmentation quality assessment. Cardiac structures include the myocardium (Myo), left and right atria (LA \& RA), left and right ventricles (LV \& RV), aorta (Ao), and pulmonary artery (PA), were segmented using a specialized TotalSegmentator model trained on sub-millimeter resolution data. For the inferior vena cava (IVC), superior vena cava (SVC), and pulmonary veins (PV), we retain the labels from the original dataset. This results in 11 channels per segmentation. To ensure anatomical validity, we perform topological filtration on all structures except the pulmonary veins, where we extract only the largest connected component. The resulting segmentations are standardized by resampling to a uniform voxel resolution of $2 \text{mm}$ and subsequently cropped to a fixed range. The crop center is determined from the union of all four chamber segmentations, and the crop length is set to 128 voxels for each side.

\noindent\textbf{Aortic Dataset} For the aorta dataset, we extract labels directly from the original TotalSegmentator v2~\cite{wasserthal2022totalsegmentator} segmentations, without applying a specialized model, resulting in 450 3D segmentations manually selected based on segmentation quality assessment. The labels include the main aortic trunk and the ascending branches, which comprise the brachiocephalic trunk (BCT), left common carotid artery (LCCA), right common carotid artery (RCCA), left subclavian artery (LSCA), and right subclavian artery (RSCA), for a total of 7 channels per segmentation. All segmentations are resampled to an isotropic voxel size of $2\,\text{mm}$ and cropped to a spatial size of $128^3$ using a crop center determined from the center of all combined tissues.

\noindent\textbf{Spinal Dataset} For the spinal dataset, we utilize the CTSpine1K dataset~\cite{deng2021ctspine1k} and extract all vertebral body segmentations, resulting in 784 3D segmentations. The segmentations include 7 cervical vertebrae (C1--C7), 12 thoracic vertebrae (T1--T12), and 5 lumbar vertebrae (L1--L5), for a total of 25 channels per segmentation. To ensure spatial consistency and anatomical completeness, all segmentations are first resampled to an isotropic voxel spacing of $1\,\mathrm{mm}$. The center of the crop box is determined from the union (voxelwise sum) of all vertebral structures in each scan, and a fixed crop of $128^3$ voxels is applied for each case.

\noindent\textbf{Coronary Dataset} For the coronary dataset, we extract coronary artery-related labels from the DISRUPT-CAD dataset~\cite{visinoni2024coronary}, consisting of 120 patients with approximately 375 OCT frames in the longitudinal (z) direction. The segmentations include lumen (Lu), calcium (Ca), and vessel wall (Ve), for a total of 4 channels per segmentation. Training samples are generated by resampling the x and y directions to $128 \times 128$ pixels while preserving the original z resolution, then randomly cropping $128$ consecutive frames along the z-axis from each patient scan. This yields approximately 360 unique 3D segmentations of size $128^3$ with an isotropic in-plane voxel spacing of approximately $0.1\,\mathrm{mm}$.

\subsection{Tasks}

\noindent\textbf{Geometric Control Tasks} We detail the task-specific hyperparameters and configurations for the geometric control tasks in \cref{tab:geom_task_configs}.
\begin{table}[t]
    \centering
    \caption{Task-specific hyperparameters and configurations for geometric control tasks.}
    \label{tab:geom_task_configs}
    \resizebox{\columnwidth}{!}{
    \begin{tabular}{l|cccc}
    \toprule
    \multicolumn{1}{c|}{} & \multicolumn{4}{c}{\textbf{Geometric Control Task}} \\
    \hline
    \textbf{Parameter} & \textbf{RV} & \textbf{Mitral} & \textbf{Aortic} & \textbf{Myo} \\
    \midrule
    Domain & Cartesian & Interface & Curvilinear & Spherical \\
    Selection Vector & [RV] & [LV], [LA] & [Ao] & [Myo] \\
    Num. Substructures & 1 & 2 & 5 & 4 \\
    \midrule
    Grid Resolution & [64,64,64] & [4,32,32] & [1,32,32] & [4,4,16] \\
    \midrule
    Mass Threshold & $10^{-5}$ & $10^{-4}$ & $10^{-6}$ & $10^{-6}$ \\
    \midrule
    $\lambda_{\mathrm{geo}}$ & 1 & 1 & 1 & 1 \\
    $\lambda_0$ (Mass) & $10^7$ & $10^9$ & $10^9$ & $10^9$ \\
    $\lambda_1$ (Centroid) & $10^5$ & $10^6$ & $10^5$ & $10^5$ \\
    $\lambda_2$ (Covariance) & $10^4$ & $10^4$ & $10^3$ & $10^4$ \\
    \bottomrule
    \end{tabular}
    }
    \end{table}

\noindent\textbf{Topological Control Tasks} We detail the task-specific hyperparameters and configurations for the topological control tasks in \cref{tab:topo_task_configs}.
\begin{table}[t]
    \centering
    \caption{Task-specific hyperparameters and configurations for topological control tasks.}
    \label{tab:topo_task_configs}
    \resizebox{\columnwidth}{!}{
    \begin{tabular}{l|cccc}
    \toprule
    \multicolumn{1}{c|}{} & \multicolumn{4}{c}{\textbf{Topological Control Task}} \\
    \hline
    \multirow{2}{*}{\textbf{Parameter}} & \textbf{Atrial} & \textbf{Branch} & \textbf{Vert.} & \textbf{Calcium} \\
    & \textbf{Separation} & \textbf{Connectivity} & \textbf{Connectivity} & \textbf{Count} \\
    \midrule
    Domain & Global & Global & Global & Global \\
    Selection Vector & [LA, RA] & All Tissues & [T6--T10] & [Ca] \\
    Num. Substructures & 1 & 1 & 1 & 1 \\
    \midrule
    Grid Resolution & [64,64,64] & [64,64,64] & [64,64,64] & [64,64,64] \\
    \midrule
    Softmax Value & 4 & 4 & 4 & 4 \\
    $\lambda_{\text{topo}}$ & 5 & 1 & 5 & 50 \\
    \midrule
    Prior $\mathbf{B}0$ & 2 & 1 & 1 & 2 \\
    Prior $\mathbf{B}1$ & 0 & 0 & 9 & 0 \\
    Prior $\mathbf{B}2$ & 0 & 0 & 0 & 0 \\
    \bottomrule
    \end{tabular}
    }
    \end{table}

\noindent\textbf{Multiscale Control}
We detail the task-specific hyperparameters and configurations for the multiscale control tasks in \cref{tab:multiscale_task_configs}. For the spinal task, we achieve multiscale control by changing the selection vector to include fewer or more vertebral bodies. For all other tasks, we change the control domain grid resolution along specified axes.
\begin{table}[t]
    \centering
    \caption{Task-specific hyperparameters and configurations for multiscale control tasks. Hyperparameters marked with a slash $/$ indicate smaller and larger domain configurations, respectively.}
    \label{tab:multiscale_task_configs}
    \resizebox{\columnwidth}{!}{
    \begin{tabular}{l|cccc}
    \toprule
    \textbf{Parameter} & \textbf{Spinal} & \textbf{Aorta} & \textbf{Myo Wall} & \textbf{Vessel Wall} \\
    \midrule
    Domain & Cartesian & Curvilinear & Spherical & Cylindrical \\
    Selection Vector & [T5--T10]/[T6--T8] & [Ao] & [Myo] & [Ca, Ve] \\
    Num. Substructures & 1 & 5 & 16 & 16 \\
    \midrule
    Grid Resolution & [64,64,64] & [1,32,32] & [1,32,32] & [1,32,32] \\
    \midrule
    Mass Threshold & $10^{-4}$ & $10^{-6}$ & $10^{-6}$ & $10^{-6}$ \\
    \midrule
    $\lambda_{\text{geo}}$ & 1 & 1 & 1 & 1 \\
    $\lambda_0$ (Mass) & $10^7$ & $10^9$ & $10^9$ & $10^9$ \\
    $\lambda_1$ (Centroid) & $10^5$ & $10^5$ & $10^5$ & $10^5$ \\
    $\lambda_2$ (Covariance) & $10^4$ & $10^3$ & $10^4$ & $10^4$ \\
    \midrule
    Domain Grid & [64,64,64] & [1,16,16]/[16,16,16] & [4,4,16]/[8,8,16] & [4,4,32]/[16,16,32] \\
    \midrule
    \bottomrule
    \end{tabular}
    }
    \end{table}

\noindent\textbf{Partial Decoding}
For the partial decoding experiments, we used a Cartesian domain with different resolutions. For Anatomica-L, both coarse and local L-parsing used grid resolutions of $[32,32,32]$, $[64,64,64]$, and $[128,128,128]$ for low, medium, and high resolutions respectively. For Anatomica-V, we used global decoding with a fixed resolution of $[128,128,128]$. We measured speed in terms of the maximum number of label maps sampled per second using the maximum allowable batch size on a single GPU. We used an A100 with 40 GB of memory for benchmarking. For geometric guidance, the wall clock time was approximately 50 seconds per sample for the highest decoding resolution with a convolutional decoder.
\subsection{Evaluation}
\noindent\textbf{Frechet Morphological Distance} To compute the morphological features, the features are normalized by the mean and standard deviation of the real data. 

\noindent\textbf{Pointcloud evaluation metrics:} To compute the point cloud metrics, we calculate NNA for every tissue label using 256 points sampled using farthest point sampling. The metric is then averaged over the number of components. To compute the pointcloud distances, we approximate Earth Mover’s Distance (EMD) through the Sinkhorn divergence \cite{feydy2019interpolating}.

\noindent\textbf{Topological Precision} To compute the Betti numbers, we take the argmax of the predicted segmentation and compute persistent homology. For a binary segmentation, the barcodes are 1 or 0 depending on the existence of the structure. We then take the sum of barcodes per dimension as the Betti number. The topological precision is then the fraction of samples with the correct Betti number per dimension.

\section{Ablation Studies}
\label{sec:supp_ablations}
\subsection{Geometric Guidance Ablations}
We aim to study the influence of individual geometric loss weightings on the geometric fidelity and generation quality. We specifically examine the influence of \textit{disentangled} geometric guidance, where, for example, we only constrain the centroid but let size and shape free to vary. To do this, we sweep over the composite geometric loss weighting $\lambda_{geo}$ for all tasks, and apply different combinations of loss weightings [$\lambda_0, \lambda_1, \lambda_2$] to activate or deactivate different geometric loss terms (see \cref{tab:geo_guidance_ablation_params}). We sample 128 samples for each experiment, with 100 sampling steps.

\begin{figure*}[h]
    \centering
    \includegraphics[width=\textwidth]{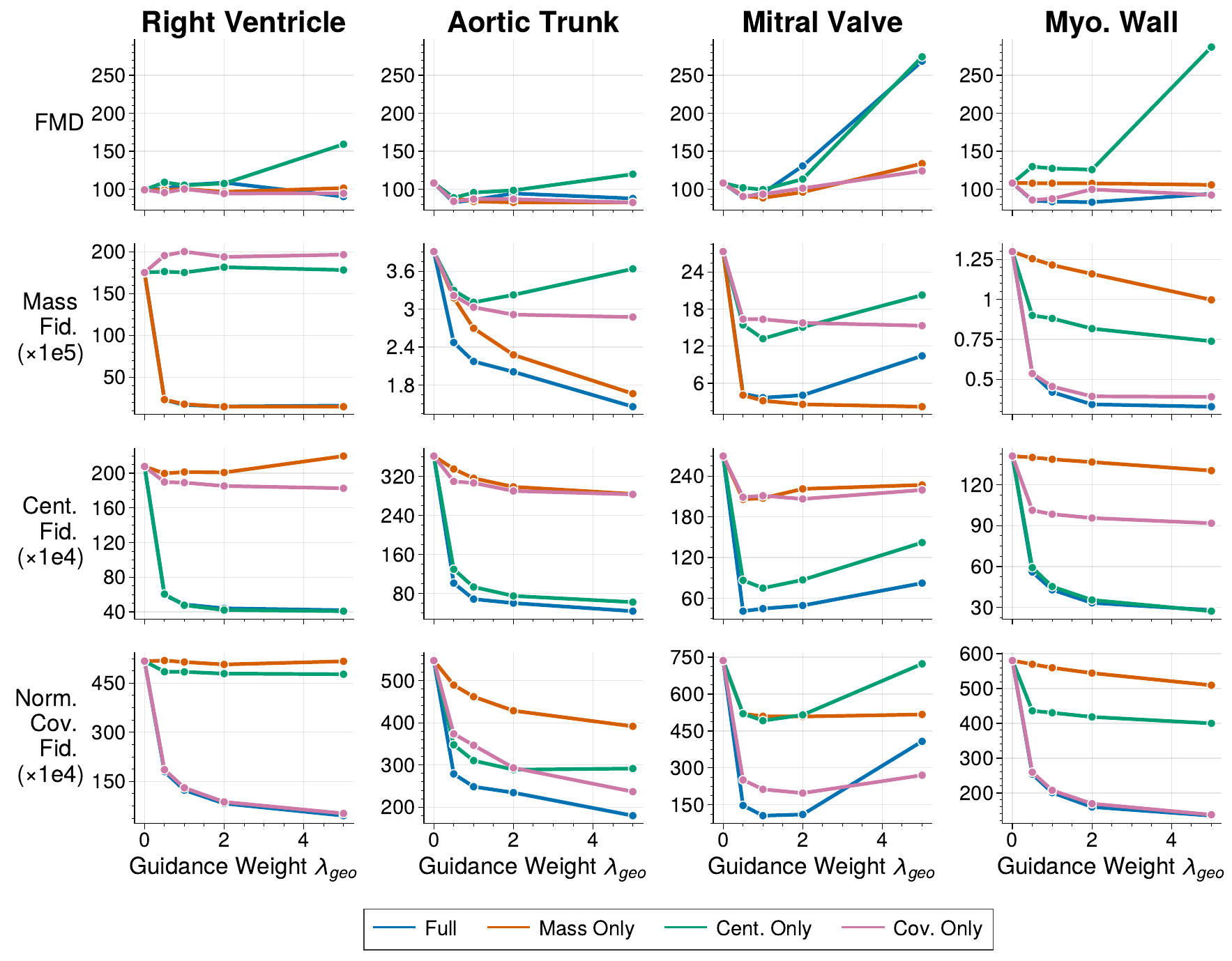}
    \caption{
        \textbf{Geometric guidance and disentangled guidance ablation study.}
    }
    \label{fig:geo_guidance_sweep}
\end{figure*}

\noindent\textbf{Effect of Guidance Weight} In \cref{fig:geo_guidance_sweep}, we see that increasing geometric guidance weight when all loss weightings are activated (Full) improves geometric fidelity up to a certain weight, after which sample quality degrades, decreasing geometric fidelity. This is especially pronounced in the case of centroid-only guidance for the mitral valve and myocardium wall tasks. For generation quality, we see similar trends where increasing guidance weights can reduce FMD up to a certain guidance weight.

\noindent\textbf{Effect of Disentangled Guidance} In \cref{fig:geo_guidance_sweep}, we demonstrate that our framework supports disentangled geometric guidance across all tasks. For instance, centroid-only guidance achieves centroid fidelity comparable to full guidance, without significantly affecting mass fidelity, shape fidelity, or generation quality as measured by FMD.

\begin{table}[h]
  \centering
  \caption{Loss weight configurations for geometric guidance ablation study.}
  \label{tab:geo_guidance_ablation_params}
  \begin{tabular}{c@{\hspace{1em}}ccc}
    \toprule
    \textbf{Guidance Loss} & $\lambda_0$ & $\lambda_1$ & $\lambda_2$ \\
    \midrule
    Full & \textcolor{green}{$\checkmark$} & \textcolor{green}{$\checkmark$} & \textcolor{green}{$\checkmark$} \\
    Mass Only & \textcolor{green}{$\checkmark$} & \textcolor{red}{$\times$} & \textcolor{red}{$\times$} \\
    Centroid Only & \textcolor{red}{$\times$} & \textcolor{green}{$\checkmark$} & \textcolor{red}{$\times$} \\
    Covariance Only & \textcolor{red}{$\times$} & \textcolor{red}{$\times$} & \textcolor{green}{$\checkmark$} \\
    \bottomrule
  \end{tabular}
\end{table}

\subsection{Topological Guidance Ablations}
\begin{figure*}[h]
    \centering
    \includegraphics[width=\textwidth]{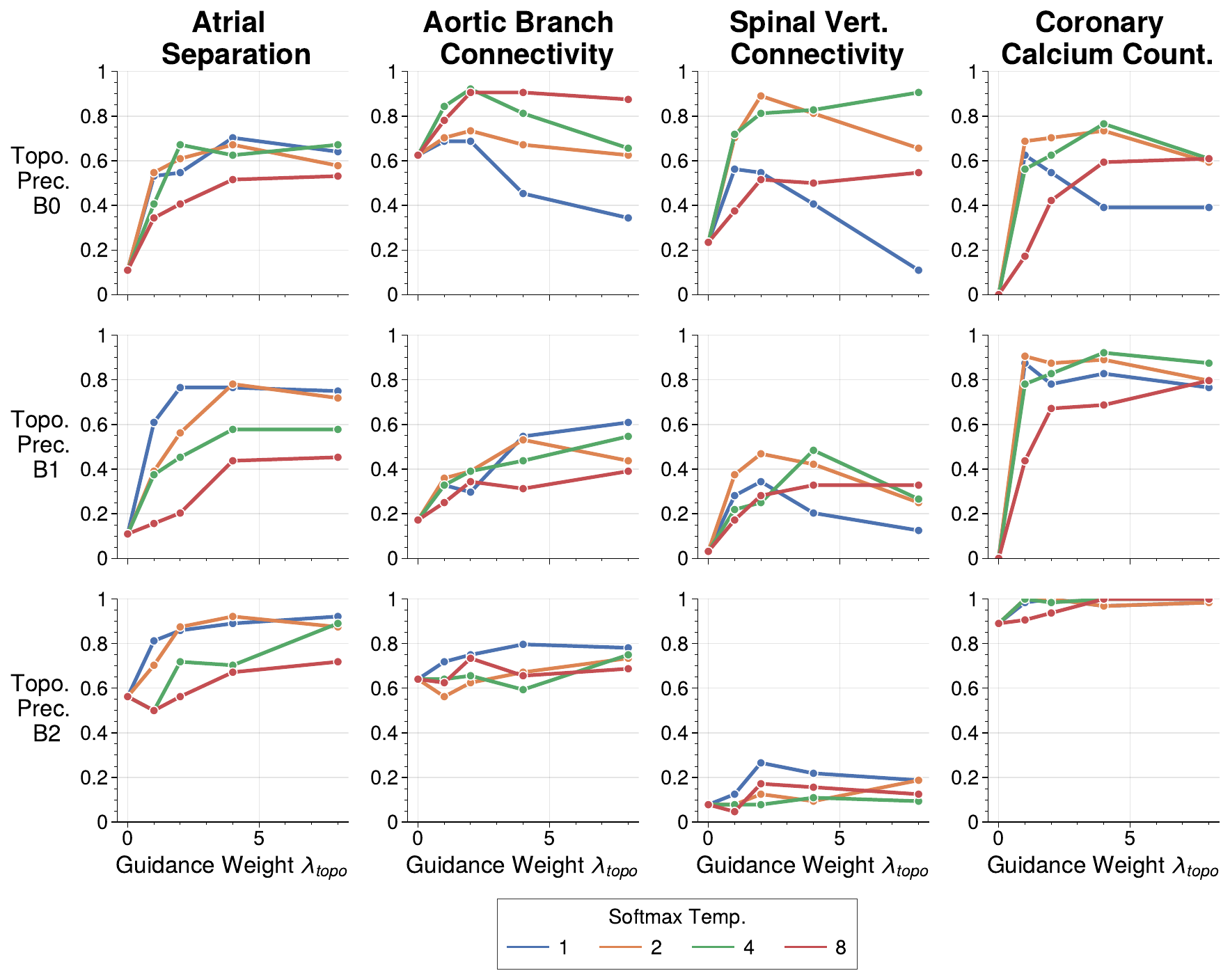}
    \caption{
        \textbf{Topological guidance and softmax temperature ablation study.}
    }
    \label{fig:topo_guidance_sweep_softmax}
\end{figure*}

\begin{figure*}[h]
    \centering
    \includegraphics[width=\textwidth]{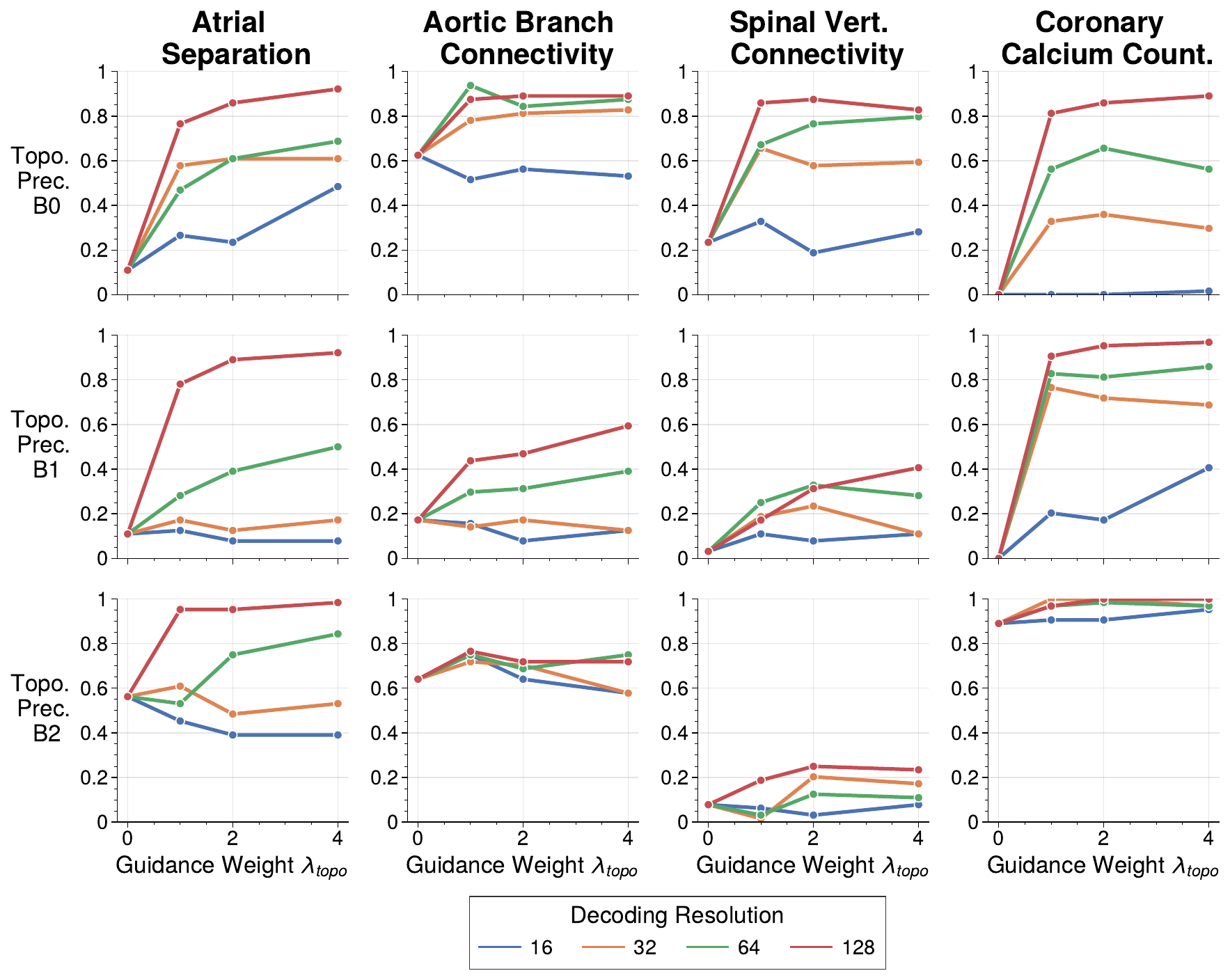}
    \caption{
        \textbf{Topological guidance and partial decoding resolution ablation study.}
    }
    \label{fig:topo_guidance_sweep_partial}
\end{figure*}

We aim to study the influence of topological loss weightings, softmax temperature, and partial decoding strategy on topological fidelity. We first sample 64 segmentations for several combinations of guidance weight and softmax temperature and evaluate topological fidelity for every combination (\cref{fig:topo_guidance_sweep_softmax}). We then sample 128 samples for various coarse decoding resolutions and guidance weights while evaluating topological fidelity (\cref{fig:topo_guidance_sweep_partial}) and sampling speed (\cref{tab:topo_speeds}).

\noindent\textbf{Effect of Guidance Weight} We see in \cref{fig:topo_guidance_sweep_softmax} that increasing guidance weights broadly improves topological fidelity but can decrease fidelity with extreme guidance weights. 

\noindent\textbf{Effect of Softmax Temperature} Similarly, in \cref{fig:topo_guidance_sweep_softmax}, we see that increasing softmax temperature can improve topological fidelity for the same guidance weight, but also improves robustness against the negative effects of exceedingly high guidance weights. The atrial separation task is an exception to this, where topological precision for loops and voids is maximized by using a softmax temperature of 1.

\noindent\textbf{Effect of Partial Decoding Strategy} We see in \cref{fig:topo_guidance_sweep_partial} that applying partial decoding with increased resolution can significantly improve topological fidelity at an increased computational cost. We find that the benefits of increased decoding resolution vary based on the topological feature and task. For example, the number of extra loops in the atrial separation task is minimized at a decoding resolution of 128, while the number of extra components for the aortic branch task is invariant after a decoding resolution of 32. We also see from \cref{tab:topo_speeds} that a decoding resolution of 64 represents a good trade-off between computational cost and topological fidelity, providing a speedup of 11x over the next highest resolution. For topological guidance, the wall clock time was approximately 420 seconds per sample for the highest decoding resolution with a convolutional decoder.

\begin{table}[h]
    \centering
    \caption{\textbf{Topological sampling speed comparison for partial decoding strategies}. Speed is measured in terms of sampled label maps per second using the maximum allowable batch size on a single GPU, normalized to the slowest method.}
    \label{tab:topo_speeds}
    \resizebox{0.6\columnwidth}{!}{\begin{tabular}{lllc}
    \toprule
    \multicolumn{3}{c}{\textbf{Methodology}} &  \\
    \cmidrule(r){1-3} \cmidrule(l){4-4}
    \textbf{Approach} & \textbf{Domain} & \textbf{Res.} & \textbf{Speed (↑)} \\
    \midrule
    \multirow{4}{*}{Anatomica-L} & \multirow{4}{*}{Coarse} & 16 & \textbf{32.00} \\
     & & 32 & 26.25 \\
     & & 64 & 11.00 \\
     & & 128 & 1.14 \\
    \midrule
    Anatomica-V & Global & 128 & 1.00 \\
    \midrule
    \end{tabular}
    }
    \end{table}




\end{document}